\documentclass[10pt,journal,compsoc]{IEEEtran}
\IEEEoverridecommandlockouts
\usepackage{amsmath,amssymb,amsfonts,bm,amsfonts}
\usepackage{algorithmic}
\usepackage{graphicx}
\usepackage{balance}
\usepackage{textcomp}
\usepackage{newfloat}
\usepackage{listings}
\usepackage[table]{xcolor}
\usepackage{algorithm}
\usepackage{multirow}
\usepackage{url}
\usepackage{booktabs}
\usepackage{colortbl,bm,amsfonts}
\usepackage{amsmath}
\usepackage{subcaption}
\usepackage{array}
\usepackage{enumitem}
\usepackage{caption}
\usepackage{tabularx}
\usepackage[T1]{fontenc}
\def\eg{\emph{e.g}.} 
\def\ie{\emph{i.e}.}

\newcommand{\fix}[1]{\textcolor{black}{{#1}}}
\usepackage{color}

\def\BibTeX{{\rm B\kern-.05em{\sc i\kern-.025em b}\kern-.08em
    T\kern-.1667em\lower.7ex\hbox{E}\kern-.125emX}}
\begin{document}
\def\method{GALA}
\definecolor{LightCyan}{rgb}{0.88,1,1}
\newcommand{\paratitle}[1]{\vspace{1ex}\noindent\emph{\textbf{#1}}}

\title{\method{}: Graph Diffusion-based Alignment with Jigsaw for Source-free Domain Adaptation}

\author{Junyu Luo, Yiyang Gu, Xiao Luo, Wei Ju, Zhiping Xiao, Yusheng Zhao, Jingyang Yuan, Ming Zhang
        
\IEEEcompsocitemizethanks{
\IEEEcompsocthanksitem Corresponding authors: Xiao Luo, Wei Ju, Zhiping Xiao.
\IEEEcompsocthanksitem Junyu Luo, Yiyang Gu, Wei Ju, Yusheng Zhao, Jingyang Yuan, and Ming Zhang are with School of Computer Science, National Key Laboratory for Multimedia Information Processing, Peking University-Anker Embodied AI Lab, Peking University, Beijing, China. (e-mail: luojunyu@stu.pku.edu.cn, yiyanggu@pku.edu.cn, juwei@pku.edu.cn, yusheng.zhao@stu.pku.edu.cn, yuanjy@pku.edu.cn, mzhang\_cs@pku.edu.cn)
\IEEEcompsocthanksitem Xiao Luo and Zhiping Xiao are with Department of Computer Science, University of California, Los Angeles, USA. (e-mail: xiaoluo@cs.ucla.edu, patricia.xiao@cs.ucla.edu)
\IEEEcompsocthanksitem This paper is partially supported by the National Key Research and Development Program of China with Grant No. 2023YFC3341203 as well as the National Natural Science Foundation of China with Grant Numbers 62276002 and 62306014.
}
}

\IEEEtitleabstractindextext{%
\begin{abstract}

Source-free domain adaptation is a crucial machine learning topic, as it contains numerous applications in the real world, particularly with respect to data privacy. Existing approaches predominantly focus on Euclidean data, such as images and videos, while the exploration of non-Euclidean graph data remains scarce. Recent graph neural network~(GNN) approaches can suffer from serious performance decline due to domain shift and label scarcity in source-free adaptation scenarios. In this study, we propose a novel method named \underline{G}raph Diffusion-b\underline{a}sed A\underline{l}ignment with Jigs\underline{a}w~(\method{}), tailored for source-free graph domain adaptation. To achieve domain alignment, \method{} employs a graph diffusion model to reconstruct source-style graphs from target data. Specifically, a score-based graph diffusion model is trained using source graphs to learn the generative source styles. 
Then, we introduce perturbations to target graphs via a stochastic differential equation instead of sampling from a prior, followed by the reverse process to reconstruct source-style graphs. We feed the source-style graphs into an off-the-shelf GNN and introduce class-specific thresholds with curriculum learning, which can generate accurate and unbiased pseudo-labels for target graphs. Moreover, we develop a simple yet effective graph-mixing strategy named graph jigsaw to combine confident graphs and unconfident graphs, which can enhance generalization capabilities and robustness via consistency learning. Extensive experiments on benchmark datasets validate the effectiveness of \method{}. The source code is available at t\url{https://github.com/luo-junyu/GALA}.

\end{abstract}

}
\maketitle

\section{Introduction}

Graph neural networks~(GNNs) have recently demonstrated superior performance in graph-level representation learning, which facilitates various graph machine learning problems. Among them, graph classification aims to predict the labels of the whole graphs~\cite{zhang2018end, ying2018hierarchical,wu2020comprehensive, pami-graphcls}, which has various real-world applications, such as molecule property prediction~\cite{lu2019molecular, kim2023learning} and social network analysis~\cite{yoo2022model,ju2022kgnn}. Typically, these methods adopt the message-passing mechanism to generate informative node-level representations~\cite{kipf2017semi}. These representations are subsequently summarized into graph-level representations via summarization operators for classification. 

In spite of the promising performance of GNN approaches, they typically operate under the assumption that training and testing graphs come from the same distribution. However, in real-world applications, this is rarely the case. 
For example, in the discovery of novel molecules in unexplored environments, particularly in response to urgent events, there is often a substantial domain shift from existing training data, resulting in inferior testing performance. One potential solution is to use graph domain adaptation methods, which combine GNN with domain alignment techniques. These methods usually utilize adversarial learning~\cite{yin2022deal} to implicitly minimize distribution differences, or use cross-domain graph contrastive learning~\cite{coco} for explicit alignment.

However, data privacy has become an important issue in both daily life and scientific research~\cite{tan2023federated}. This is evident in the practices of AI companies, which generally release only pre-trained models rather than the entire dataset. Recent graph domain adaptation approaches typically require complete access to both the source and target graphs~\cite{lin2023multi, coco}, which can be impracticable due to privacy regulations. Towards this end, this paper investigates source-free graph domain adaptation, a practical but under-explored problem, which adapts an off-the-shelf source model to unlabeled target graphs without accessing the source graphs. In addition, our scheme is data-efficient, as we sufficiently leverages the open-source pre-trained models. 

In literature, various works have focused on source-free domain adaptation~\cite{sun2020test, shot, nado2020evaluating, gao2023back, plue}. These strategies usually employ self-supervised techniques for semantic exploration in the target domains and conduct domain alignment based on training statistics and weight matrices.
However, these works often concentrate on normal Euclidean data, such as images and text, while this problem remains under-explored on non-Euclidean graph data. In fact, developing a source-free graph domain adaptation framework meets two significant challenges. 
(1) \textit{How to sufficiently align two graph domains without having access to the source graphs?} 
\fix{The graph samples contain complicated structures, including various nodes and topologies~\cite{velickovic2019deep}. The complicated structures and rich semantics lead to a hierarchical domain shift, which is difficult to capture~\cite{you2022graph}.} Even worse, we are unable to explicitly align two domains by minimizing the distribution divergence, since source graphs are unavailable. (2) \textit{How to overcome the label scarcity in the target domain?} \fix{In real-world applications, we can only access limited unlabeled target graphs, since labeling graph-structured data can be very costly~\cite{pami-rev-2}. Previous techniques often adopt pseudo-labeling for additional supervision. Given the complexities inherent in graph structures, these pseudo-labels probably contain errors and are skewed toward the dominant class. }

\begin{figure}[t]
    \centering
    \includegraphics[width=\linewidth]{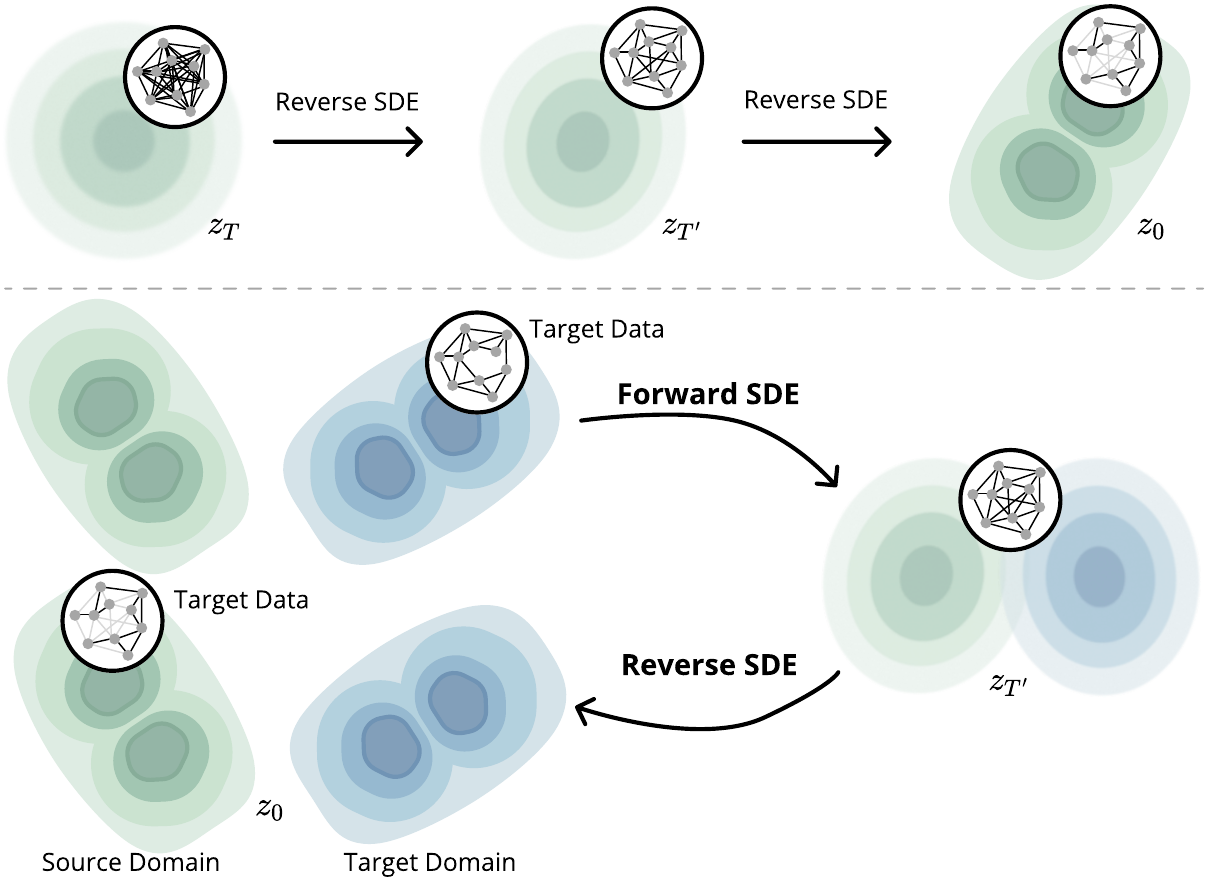}
    \caption{A brief motivation of \method{}. Previous score-based diffusion models usually generate new data by sampling from a prior distribution~(upper), while ours \method{} transforms the target graphs back to the source domain~(lower). }
    \label{fig:fig1}
\end{figure}

To tackle these two challenges, this paper proposes a novel method named \underline{G}raph Diffusion-b\underline{a}sed A\underline{l}ignment with Jigs\underline{a}w (\method{}), tailored for source-free graph domain adaptation. The main idea of our \method{} is to convert the target graphs into source-style graphs using a graph diffusion model. Specifically, we first employ source graphs to train a score-based graph diffusion model, which connects intricate source graph structures with a prior distribution using a stochastic differential equation~(SDE). 
As shown in Figure~\ref{fig:fig1}, we use standard diffusion models to generate new data from a prior distribution, while we perturb target graphs using forward SDE, and then reconstruct source-style graphs using reverse SDE with source styles. 
In this way, we bridge target graphs with the source domain in the protection of data privacy. 

In addition, to generate more accurate and impartial pseudo-labels under label scarcity, we introduce adaptive class-specific thresholds, which will progressively increase with the spirit of curriculum learning. Target graphs are regarded as confident if their confidence scores are above the threshold. To exploit unconfident graphs, we provide a simple yet effective graph-mixing method called graph jigsaw, which executes graph clustering based on community detection to select a subgraph from each graph. 
Subsequently, a confident graph and an unconfident graph exchange subgraphs, much like a jigsaw puzzle. Mixed and original graphs are assumed to have consistent predictions for model generalization and robustness. We conduct comprehensive experiments on benchmark datasets and experiments demonstrate the superiority of the proposed \method{} compared with existing baselines. 

To summarize, the contribution of this work is as follows:
\begin{itemize}
\item We explore a practical but under-explored problem of source-free graph domain adaptation. To the best of our knowledge, we are the first to explore this problem. 
\item We propose a novel framework named \method{}, which transforms target graphs into source-style using a graph diffusion model to mitigate the domain shift in data. 
\item To overcome the label scarcity, we not only introduce adaptive class-specific thresholds with curriculum learning to learn accurate and unbiased pseudo-labels, but also utilize graph jigsaw with consistency learning for model generalization and robustness. 
\item Extensive experiments on a range of benchmark datasets demonstrate the superiority of the proposed \method{} compared with various competing baselines. Extensive ablation studies and visualization further validate our superiority. 

\end{itemize}

\section{Related Work}

In this section, we briefly review three related topics, \ie, graph classification, graph domain adaptation, and source-free domain adaptation. 

\subsection{Graph Classification}

Graph neural networks (GNNs) have remarkable performance in graph classification by mapping graph-structured data into embedding vectors~\cite{wu2020comprehensive, luo2022dualgraph, pami-graphcls,ju2024comprehensive}. Among various GNNs, message passing neural networks have been the most prevalent~\cite{kipf2017semi,xu2019powerful}, which update node representations via neighborhood aggregation. Recently, graph kernels have been incorporated into GNNs to learn from various substructures such as motifs and paths~\cite{cosmo2021graph}. 
To generate graph-level representations, various graph pooling operators are proposed using the attention mechanism~\cite{lee2019self}, reinforcement learning~\cite{sun2021sugar}, and graph clustering~\cite{bianchi2020spectral}. In addition, various semi-supervised methods~\cite{li2019semi} are proposed for data-efficient graph-level learning~\cite{ju2022glcc}. Despite extensive progress, existing methods mostly assume that training and test graphs come from the same distribution, which is usually not the case in real-world applications. Therefore, we adopt diffusion adaptation to transform target graphs into source-style graphs effectively.

\subsection{Graph Domain Adaptation} 

Graph domain adaptation has attracted substantial interest in recent research~\cite{long2015learning, lee2013pseudo, assran2021semi, zhang2021flexmatch, pami-semisup, pami-selfsup, pami-selfsup2, pami-unsup, pami-unsup2}. Early efforts typically focus on node-level adaptation~\cite{uda-gcn, wu2021handling, liu2022confidence, lin2023multi}, where knowledge is transferred from a labeled source graph to an unlabeled target graph. These strategies usually utilize improved GNN encoders and adversarial learning for domain-invariant representations. 
For instance, UDA-GCN~\cite{uda-gcn} uses dual graph encoders to explore semantics from multiple perspectives and then combines them with the attention mechanism. 
In contrast to node-level, graph-level adaptation would encounter numerous graphs with complex distribution shifts~\cite{wu2021discovering, buffelli2022sizeshiftreg}. 
The recent CoCo~\cite{coco} combines graph kernel networks with GNNs and employs a coupled graph contrastive learning framework to tackle domain disparities. 
\fix{StruRW~\cite{pami-rev-1} utilizes a structural reweighting method to address the domain shift caused by graph structure and node attributes.
SA-GDA~\cite{pami-rev-2} uses spectral augmentation and dual graph convolutional networks for graph domain adaptation.
UDANE~\cite{pami-rev-3} leverages the transferable node representation to transfer knowledge across domains.
P-Mixup~\cite{pami-rev-4} enforces the target domain progressively moving to the source domain for effective domain transfer.}
However, these works generally require access to both source and target graphs, which could be impractical in real-world scenarios with strict privacy restrictions. To this end, we investigate source-free graph domain adaptation, which adapts an off-the-shelf model to an unlabeled target domain.

\subsection{Source-Free Domain Adaptation}

Source-free domain adaptation seeks to transfer a model trained on a source domain to an unlabeled target domain without access to source data. Existing approaches can be essentially categorized into self-training approaches and domain alignment approaches. Self-training approaches typically employ pseudo-labeling and mutual information maximization to discover semantics from unlabeled target data~\cite{sun2020test, saito2017asymmetric, shot}. 
For instance, SHOT~\cite{shot} generates prototypes of classes in the hidden space, which can be used to train the nearest centroid classifier for unlabeled target data. 
In contrast, domain alignment approaches attempt to explore stored batch statistics and weight matrices for reduced distribution shift~\cite{nado2020evaluating, Schneider_Rusak_Eck_Bringmann_Brendel_Bethge_2020}. However, these methods are designed for visual data, which performs poorly on non-Euclidean graph data. Towards this end, we propose a diffusion-based source-free graph domain adaptation method, which transforms target graphs back to the source domain via a graph diffusion model. 

\begin{figure*}[t]
    \centering
    \includegraphics[width=\textwidth]{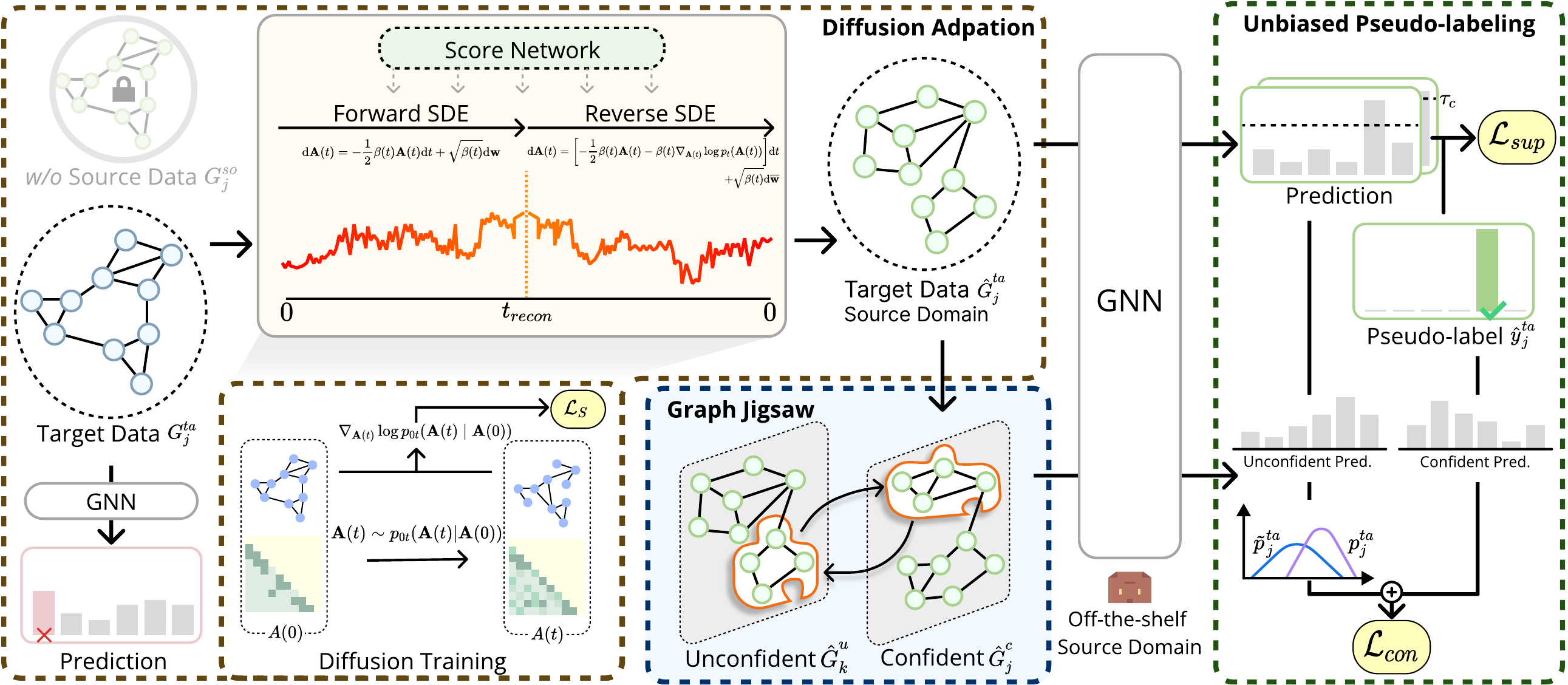}
    \caption{Overview of \method{}. We employ source graphs to train a score-based graph diffusion model and then transform target graphs into source-style graphs. Moreover, we introduce adaptive class-specific thresholds to generate confident graphs with pseudo-labels and then utilize graph jigsaw to exchange subgraphs between graph pairs for consistency learning. }
    \label{fig:fig2}
\end{figure*}

\section{Preliminaries}

\subsection{Problem Definition}
We denote a graph as $G = (V, E)$ where $V$ denotes nodes and $E$ denotes edges. Each graph is associated with a node attribute matrix $\bm{X} \in \mathbb{R}^{|V| \times d^f}$ where $d^f$ is the attribute dimension. We denote a labeled source domain as $\mathcal{D}^{so} = \{(G_i^{so}, y_i^{so})\}_{i=1}^{N_s}$ where $G_i^{so}$ stands for the $i$-th source graph and $y_i^{so}$ is its label. An unlabeled target domain is $\mathcal{D}^{ta} = \{G_j^{ta}\}_{j=1}^{N_t}$ where $G_j^{ta}$ denote the $j$-th target graph. While two domains share the same label space, they have different data distributions. In the problem of source-free graph domain adaptation, we initially pre-train neural network models using source data and subsequently learn a model to predict the labels of target graphs. Importantly, to safeguard data privacy, the source data remains inaccessible during the adaptation on the target domain.

\subsection{Graph Neural Networks}
Graph neural networks~(GNNs) are commonly used to encode graph-structured data based on a message-passing mechanism. In particular, given a graph $G$, we denote the node representation of $v\in V$ at the layer $k$ as $\bm{h}_i^{(k)}$ and formulate the updating rule as follows:
\begin{equation}
    \bm{h}_{\mathcal{S}(v)}^{(k)} =\mathrm{AGG}^{(k)}\left(\left\{\bm{h}_{u}^{(k-1)}: u \in \mathcal{S}(v)\right\}\right),
\end{equation}
\begin{equation}
    \bm{h}_{v}^{(k)} =\mathrm{COM}^{(k)}(\bm{h}_{v}^{(k-1)}, \bm{v}_{\mathcal{S}(v)}^{(k)}),
\end{equation}
where $\mathcal{S}(v)$ represents the neighboring nodes of $v$ and $\bm{h}_{\mathcal{S}(v)}^{(k)}$ denotes the neighborhood representations for $v$. $\mathrm{AGG}^{(k)}$ and $\mathrm{COM}^{(k)}$ denote the aggregation and combination operators. 
Finally, a global pooling function is used to summarize the node representations into a graph-level representation,
\begin{equation}\label{eq:gp}
\bm{z} = \mathrm{GP}\left(\left\{\bm{v}_{i}^{(K)}\right\}_{i=1}^n\right),
\end{equation}
in which $\mathrm{GP}(\cdot)$ is for global pooling. Finally, we utilize an MLP classifier with softmax activation, \ie, $\mathrm{HEAD}(\cdot)$ to generate label distributions:
\begin{equation}\label{eq:gp2}
\bm{p} = \Phi(G) = \mathrm{HEAD}(\bm{z}),
\end{equation}
where $\Phi$ represents the whole graph neural network. In our setting, we would first train an off-the-shelf GNN model using source graphs by minimizing the cross-entropy objective as a preliminary:
\begin{equation}
\mathcal{L}_{source}=-\frac{1}{|\mathcal{D}^{so}|} \sum_{G_{i}^{so} \in \mathcal{\mathcal{D}}^{so}} \log \boldsymbol{p}_{i}^{so}\left[y_{i}^{so}\right].
\end{equation}
where $\boldsymbol{p}_{i}^{so} = \Phi(G_{i}^{so})$ denotes the output of the GNN.

\section{The Proposed Approach}
\subsection{Framework Overview}

This paper introduces \method{} for source-free graph domain adaptation. The model, as illustrated in Figure~\ref{fig:fig2}, does not have access to source graphs when learning from target graphs. To solve the problem, we are required to sufficiently align two graph domains and overcome the label scarcity on the target domain. 
At a high level, \method{} transforms target graphs into source-style with a graph diffusion model.
Specifically, source graphs are utilized to learn a score-based graph diffusion model, which will be fixed to keep the source styles. 
On this basis, we perturb target graphs by forwarding a stochastic differential equation~(SDE) and then reconstruct source-style graphs using the reverse process. 
Moreover, we adapt the off-the-shelf GNN model with class-specific thresholds through curriculum learning. This aids in generating accurate and class-balanced pseudo-labels for each target graph. Then, we introduce our proposed \method{} in detail as below. 

\subsection{Graph Diffusion Model for Domain Alignment}

The principal challenge faced in graph domain adaptation is the alignment of source and target domains. 
This becomes increasingly challenging in our scenarios since we cannot simultaneously access the source and target data.
Our approach is to employ a generative model for source-style graphs that can make precise predictions using the pre-trained GNNs. To achieve this, we introduce an off-the-shelf graph diffusion model in the source domain, with the ability to generate graphs with source style. Subsequently, the target data can be fed into the diffusion model to generate a source-style graph via the reverse process. 

\subsubsection{Graph Diffusion Model} We first introduce our graph diffusion model, which reconstructs the graph structure using an SDE~\cite{song2020score, 2021SDEdit, graphgdp, pami-graphdiff}.
As the most crucial component, the graph structure is expressed via the adjacency matrix as a variable from the timestamp $0$ to $1$ with an interval $\Delta t$ when adding noise, \ie, $\mathbf{A}(t)$. Then, the SDE for the adjacency matrix is:
\begin{equation}
\mathrm{d} \mathbf{A}(t)=-\frac{1}{2} \beta(t) \mathbf{A}(t) \mathrm{d} t+\sqrt{\beta(t)} \mathrm{d} \mathbf{w},
\end{equation}
where $\beta(t)$ is a scalar of adding noise and $\mathbf{w}$ is a Wiener process indicating random noise. The key of an effective diffusion model is to obtain the score function, \ie, the gradient of the likelihood $\nabla_{\mathbf{A}({t})} \log p_{0 t}\left(\mathbf{A}({t}) \mid \mathbf{A}(0)\right)$ where $\mathbf{A}(t)\sim p_{0t} (\mathbf{A}(t)|\mathbf{A}(0))$. Here, we utilize a score-based GNN $\rho(\mathbf{A}(t),t)$ to approximate the ground truth by minimizing the following objective:
\begin{equation}\label{eq:diffusion}
\begin{aligned} 
\mathcal{L}_S= \mathbb{E}_{t}\left\{\lambda ( t ) \mathbb { E } _ { \mathbf { A }(0) } \mathbb { E } _ { \mathbf { A }(t) | \mathbf { A }(0)} \left[\| \rho \left(\mathbf{A}(t), t\right)\right.\right. \\ \left.\left.-\nabla_{\mathbf{A}(t)} \log p_{0 t}\left(\mathbf{A}(t) \mid \mathbf{A}(0)\right) \|_{2}^{2}\right]\right\},
\end{aligned}
\end{equation}
where $\lambda(t)$ is a positive weighting function. By minimizing Eq.~\ref{eq:diffusion} on the source domain, we can obtain a graph diffusion model with source styles embedded.

\subsubsection{Graph-based Score Function}
Different from diffusion models in computer vision, our approach requires incorporating graph-structural information during propagation. Combining a random walk with a message-passing mechanism, we develop an effective graph-based score function $\rho(\mathbf{A}(t),t)$. Specifically, to better learn structural information, we first discretize the adjacency matrix and then utilize a random walk to generate edge representations $\bm{e}_{mn}$:
\begin{equation}
    \dot{\mathbf{A}}(t) = \bm{1}_{\mathbf{A}(t)>1/2},
\end{equation}
\begin{equation}
    \bm{e}^{(0)}_{mn} = \phi([\bm{R}_{mn},\bm{R}^2_{mn},\cdots, \bm{R}^r_{mn} ]), 
\end{equation}
where $\bm{R}^{k} = (\dot{\mathbf{A}}(t) \dot{\mathbf{D}}^{-1}(t))^{k} $ indicates the likelihood of starting and end points for every $k$-length walk. $r$ denotes the maximal random walk length and $\dot{\mathbf{D}}^{-1}(t)$ is the degree matrix of $\dot{\mathbf{A}}(t)$. $\phi(\cdot)$ is an MLP to map these likelihoods into the embedding space.
Besides edge information, we also extract node representations $\bm{f}_i^{(0)} = [\bm{e}_{ii},\bm{d}_i]$ where $\bm{d}_i$ is the one-hot degree vector, and then utilize a message passing neural network with to update $\bm{f}_i^{k}$ as:
\begin{equation}
    \bm{f}_{\mathcal{S}(v)}^{(k)} =\mathrm{AGG}^{(k)}\left(\left\{\bm{f}_{i}^{(k-1)}: j \in \mathcal{S}(i)\right\}\right)
\end{equation}
\begin{equation}
    \bm{f}_{i}^{(k)} =\mathrm{COM}^{(k)}(\bm{f}_{i}^{(k-1)}, \bm{f}_{\mathcal{S}(v)}^{(k)}).
\end{equation}
Finally, we update edge representations as:
\begin{equation}
    \bm{e}^{(k+1)}_{ij} = \widetilde{\mathrm{COM}}^{(k)}( \bm{e}^{(k)}_{ij}, \bm{f}^{(k)}_{i}, \bm{f}^{(k)}_{j}),
\end{equation}
where $\widetilde{\mathrm{COM}}^{(k)}$ is a different combination operator to update the edge representation. After stacking $K^d$ layers, we utilize an MLP to generate the estimated adjacency matrix. 

\subsubsection{Target Graph Adaptation} By sampling from the prior distribution, the well-trained graph diffusion model can generate new graphs carrying source styles. Instead of sampling from the prior distribution $p_1(\mathbf{A}(1))$, \method{} adds noise to the target graphs and then uses the reverse process to endow the target graphs with source styles. This process only utilizes the trained diffusion model, effectively addressing the domain shift while protecting the privacy of source graphs. 

In detail, we perturb the adjacency matrix of the target graph, which generates noise with target semantics,
\begin{equation}
    \mathbf{A}(t_{recon})\sim p_{0t_{recon}}(\mathbf{A}(t_{recon})|\mathbf{A}(0)).
\end{equation}
where $t_{recon}$ is the starting point of reconstruction. Consequently, we utilize an iterative process to alter the noisy input $\mathbf{A}(t_{recon})$ for reconstruction with the reverse process. The reserve SDE can be written as:
\begin{equation}
\begin{aligned}
    \mathrm{d} \mathbf{A}(t)& =\left[-\frac{1}{2} \beta(t) \mathbf{A}(t)-\beta(t) \nabla_{\mathbf{A}(t)} \log p_{t}(\mathbf{A}(t))\right] \mathrm{d} t \\ & +\sqrt{\beta(t)} \mathrm{d} \overline{\mathbf{w}},
\end{aligned}
\end{equation}
where $\overline{\mathbf{w}}$ denotes a standard Wiener process for backward. On this basis, the updating process of the generated adjacency matrix $\mathbf{A}^g(t-\Delta t)$ can be: 
\begin{equation}
    \mathbf{A}^g(t-\Delta t)\sim p(\mathbf{A}^g(t-\Delta t)|\mathbf{A}^g(t)).
\end{equation}
In our implementation, the Euler-Maruyama algorithm is employed to simply the propagation as:
\begin{equation}
\begin{aligned}
\mathbf{A}^g({t-\Delta t})&=\mathbf{A}^g(t)+\left[\frac{1}{2} \beta(t) \mathbf{A}^g(t)+\beta(t) \mathbf{s}_{\theta}\left(\mathbf{A}^g(t), t\right)\right] \\ &+\sqrt{\beta(t)} \sqrt{\Delta t} \mathbf{z}_{t},
\end{aligned}
\end{equation}
where $\mathbf{z}_{t}\sim \mathcal{N}(0,1)$. Finally, the adjacency matrix $\mathbf{A}^g(0)$ is for the reconstructed graph, which is denoted as $\hat{G}_j^{ta}$. In this manner, we can adapt the target graphs into source-type graphs from a data-centric perspective, which can facilitate accurate semantics exploration using the out-of-shelf GNN.  



\subsection{Unbiased Pseudo-labeling with Curriculum Learning}

To overcome label scarcity in the target domain, we turn to pseudo-labeling techiniques. Note that after transforming target graphs into source-style graphs, we can use the pre-trained GNN with a fixed threshold to generate target graph pseudo-labels~\cite{sohn2020fixmatch}.
However, these pseudo-labels from the pre-trained GNN could be biased toward the dominant class. Worse still, GNNs could produce overconfident pseudo-labels due to the potential class competition from normalization operators. To avoid this, we propose unbiased pseudo-labeling with class-specific thresholds for accurate and unbiased pseudo-labels. In addition, a curriculum learning approach is employed, which initially concentrates on all graph samples and then progressively focuses on more reliable ones. In other words, we increase the threshold to reject samples from easy to hard. 

Primarily, we generate the label distribution $\bm{p}_j^{ta}$ using the pre-trained GNN for each source-style target graph $\hat{G}^{ta}_j$, \ie,
\begin{equation}
    \bm{p}_j^{ta} = \Phi(\hat{G}^{ta}_j),
\end{equation}
where $\hat{G}^{ta}_j$ is the output of the diffusion model. To generate the rigid and class-balanced pseudo-labels, we first calculate the confidence distribution for each class and subsequently establish an adaptive threshold based on the maximum. In formulation, the confidence of each sample is defined as:
\begin{equation}
    s_j^{ta} = max_{c} \bm{p}_j^{ta}[c].
\end{equation}
Then, the maximum confidence scores for class $c$ are:
\begin{equation}
    \mathcal{M}_c = \max \{s_j^{ta}| argmax_{c'} \bm{p}_j^{ta}[c'] = c  \}.
\end{equation}
The adaptive threshold for class $c$ is defined as:
\begin{equation}\label{eq:threshold}
    \tau_c = \mathcal{M}_c \cdot \alpha(e),
\end{equation}
where $\alpha(e)$ is shared across different classes.
Following the spirit of curriculum learning, we would increase $\alpha(e)$ linearly according to epoch number $e$, which gradually selects graphs with higher confidence. Consequently, this would yield a collection of confident target graphs:
\begin{equation}\label{eq:confident}
    \mathcal{C} = \{G_j^{ta}| c = argmax_{c'} \bm{p}_j^{ta}[c'] , s_j^{ta} >\tau_c,   \}.
\end{equation}
The standard cross-entropy loss is minimized in $\mathcal{C}$:
\begin{equation}
    \mathcal{L}_{sup} = - \frac{1}{|\mathcal{C}|} \sum_{G_j^{ta} \in \mathcal{C}} \log \bm{p}_j^{ta}[\hat{y}_j^{ta}],
\end{equation}
where $\hat{y}_j^{ta}$ denotes the pseudo-label of $G_j^{ta}$. Our pseudo-labeling strategy provides a reliable and unbiased optimization process with reduced error accumulation. 


\subsection{Consistency Learning with Graph Jigsaw}

Despite the effectiveness of pseudo-labeling, there could be a large number of unconfident target graphs without sufficient exploration. To tackle this, we aim to expand the dataset by combining confident and unconfident graphs. 
Recently, a number of works have been developed on graph Mixup~\cite{graphmad, han2022g}, which combines graph representations in the latent space. However, these methods cannot promise to generate realistic graphs and are dependent on GNN encoders. In contrast, we employ a simple yet effective augmentation strategy, graph jigsaw, that works in the graph space. Graph jigsaw uses a community detection algorithm to derive a small subgraph from each graph and then exchange them in a jigsaw puzzle mechanism. Finally, we propose consistency learning to ensure that post-augmentation predictions remain consistent, thereby enhancing robustness and generalization. 

In detail, we adopt a community detection algorithm, \ie, the Louvain algorithm~\cite{louvain} to partition graphs into diverse clusters. The benefit of this algorithm is unnecessary to decide the cluster number. Then, we randomly choose a cluster, from each confident graph $\hat{G}_j^c$ and denote the complementary part and subgraph as, $\hat{G}_{j,1}^c$ and $\hat{G}_{j,2}^c$. In parallel, each unconfident graph $\hat{G}_k^u$ can be segmented into $\hat{G}_{k,1}^u$ and $\hat{G}_{k,2}^u$. We exchange semantics by combining $\hat{G}_{j,1}^c$ and $\hat{G}_{k,2}^u$ to generate augmented graph $\hat{G}_j^c$, and $\hat{G}_{j,2}^c$ and $\hat{G}_{k,1}^u$ would be connected to generate graph $\tilde{G}_k^u$. A detailed example can be seen in Figure \ref{fig:fig2}. 

Then, consistency learning is leveraged to encourage the mixed graph to yield similar predictions. In formulation, the learning objective is written as:
\begin{equation}
    \mathcal{L}_{con} = - \frac{1}{|\mathcal{S}|} \sum_{G_j^{ta} \in \mathcal{S}} \log \tilde{\bm{p}}_j^{ta}[\hat{y}_j^{ta}]  - \frac{1}{|\mathcal{U}|} \sum_{G_j^{ta} \in \mathcal{U}} KL(\tilde{\bm{p}}_j^{ta}||{\bm{p}}_j^{ta}),
\end{equation}
where $\tilde{\bm{p}}_j^{ta}$ is the label distribution of the augmented view and $\mathcal{U}=\mathcal{D}^{ta}/ \mathcal{S} $ are unconfident graphs. Moreover, pseudo-labels are adopted to supervise confident samples, while the KL divergence of predictions between unconfident graphs and their augmented views are minimized. Our \method{} enjoys the regularization from consistency learning to learn from extensive unconfident graph samples, which can further minimize the information loss.

\begin{algorithm}[tb]
    \caption{Optimization Algorithm of \method{}}\label{alg1}
    \label{alg:algorithm}
    \textbf{Input}: Source graphs $\mathcal{D}^{so}$, target graphs $\mathcal{D}^{ta}$, \\
    \textbf{Output}: GNN-based classifier $\Phi(\cdot)$

    \begin{algorithmic}[1] 
    \STATE Pre-train $\Phi(\cdot)$ using source graphs;
    \STATE Train diffusion model using source graphs by minimizing Eq.~\ref{eq:diffusion};
    \FOR{epoch = 1, 2, $\cdots$}
    \STATE Obtain class-specific thresholds using Eq.~\ref{eq:threshold};
    \STATE Generate confident target graphs with pseudo-labels using Eq.~\ref{eq:confident};
    \FOR{each batch}
    \STATE Sample a mini-batch of target graphs;
    \STATE Calculate the loss objective using Eq.~\ref{final_loss};
    \STATE Update the parameters of GNN through back-propagation;
    \ENDFOR
    \ENDFOR
    \end{algorithmic}
\end{algorithm}

\subsection{Summarization}

In a nutshell, the final loss of training our \method{} is summarized as:
\begin{equation}\label{final_loss}
    \mathcal{L} = \mathcal{L}_{sup} + \mathcal{L}_{con}.
\end{equation}
As a preliminary, we adopt an off-the-shelf GNN and graph diffusion model trained on the source domain.
Subsequently, the loss objective is minimized in the target domain. The overall algorithm is summarized in Algorithm~\ref{alg1}. 

\noindent\textbf{Complexity Analysis.} We assume that $|D^{ta}|$ is the number of target graphs, $d$ is the feature dimension, $|V|^a$ is the average number of nodes, $W$ is the reconstruction steps. The computational complexity of the diffusion is $\mathcal{O}(K^d|D^{ta}||V|^a{d}^2W)$. The GNN model takes $\mathcal{O}(K|D^{ta}||V|^a{d}^2)$ where $K$ is the layer number. Therefore, the complexity of our \method{} is $\mathcal{O}((K+K^dW)|D^{ta}||V|^a{d}^2)$ which is linear to both $|V|^a$ and $|D^{ta}|$.

\section{Experiments}

\begin{table*}[tb]
\centering
\tabcolsep=2pt
\caption{The classification results (in \%) on ENZYMES (source$\rightarrow$target). E0, E1, E2, and E3 are split by the graph density.}
\resizebox{\textwidth}{!}{
\begin{tabular}{llllllllllllll}
\toprule[1pt]
{\bf Methods} &E0$\rightarrow$E1 &E1$\rightarrow$E0 &E0$\rightarrow$E2 &E2$\rightarrow$E0 &E0$\rightarrow$E3 &E3$\rightarrow$E0 &E1$\rightarrow$E2 &E2$\rightarrow$E1 &E1$\rightarrow$E3 &E3$\rightarrow$E1 &E2$\rightarrow$E3 &E3$\rightarrow$E2 &Avg.\\
\midrule  
\midrule
GCN &
33.3{\tiny $\pm 4.7$}
&
29.3{\tiny $\pm 2.7$}
&
22.0{\tiny $\pm 2.2$}
&
21.3{\tiny $\pm 2.0$}
&
27.3{\tiny $\pm 2.3$}
&
16.7{\tiny $\pm 3.7$}
&
24.7{\tiny $\pm 2.4$}
&
24.7{\tiny $\pm 3.8$}
&
20.7{\tiny $\pm 0.3$}
&
30.0{\tiny $\pm 3.7$}
&
28.7{\tiny $\pm 2.8$}
&
30.7{\tiny $\pm 1.5$}
&
25.7{\tiny $\pm 2.7$}
\\
GIN &
27.3{\tiny $\pm 2.8$}
&
22.0{\tiny $\pm 3.1$}
&
24.7{\tiny $\pm 4.0$}
&
23.3{\tiny $\pm 3.3$}
&
20.7{\tiny $\pm 3.5$}
&
18.0{\tiny $\pm 2.2$}
&
23.3{\tiny $\pm 3.4$}
&
25.3{\tiny $\pm 2.5$}
&
27.3{\tiny $\pm 2.3$}
&
30.7{\tiny $\pm 2.7$}
&
23.3{\tiny $\pm 2.2$}
&
28.0{\tiny $\pm 2.5$}
&
24.5{\tiny $\pm 2.9$}
\\
GraphSAGE &
32.7{\tiny $\pm 2.9$}
&
25.3{\tiny $\pm 0.7$}
&
21.3{\tiny $\pm 2.8$}
&
22.7{\tiny $\pm 1.9$}
&
20.0{\tiny $\pm 3.0$}
&
16.0{\tiny $\pm 3.1$}
&
24.7{\tiny $\pm 2.8$}
&
24.7{\tiny $\pm 3.1$}
&
24.0{\tiny $\pm 3.3$}
&
25.3{\tiny $\pm 2.5$}
&
26.7{\tiny $\pm 2.6$}
&
34.7{\tiny $\pm 3.6$}
&
24.8{\tiny $\pm 2.7$}
\\
GAT &
29.3{\tiny $\pm 2.1$}
&
24.0{\tiny $\pm 1.5$}
&
28.0{\tiny $\pm 3.6$}
&
23.3{\tiny $\pm 3.3$}
&
23.3{\tiny $\pm 2.7$}
&
20.7{\tiny $\pm 2.9$}
&
24.0{\tiny $\pm 2.7$}
&
27.3{\tiny $\pm 1.5$}
&
23.3{\tiny $\pm 3.0$}
&
26.7{\tiny $\pm 2.6$}
&
28.0{\tiny $\pm 2.4$}
&
23.3{\tiny $\pm 2.0$}
&
25.1{\tiny $\pm 2.5$}
\\
Mean-Teacher &
16.0{\tiny $\pm 2.3$}
&
22.0{\tiny $\pm 3.6$}
&
21.3{\tiny $\pm 3.3$}
&
20.0{\tiny $\pm 2.7$}
&
20.0{\tiny $\pm 3.2$}
&
20.0{\tiny $\pm 3.7$}
&
26.7{\tiny $\pm 2.2$}
&
19.3{\tiny $\pm 1.5$}
&
18.7{\tiny $\pm 3.8$}
&
18.7{\tiny $\pm 2.0$}
&
23.3{\tiny $\pm 3.4$}
&
22.0{\tiny $\pm 2.5$}
&
20.7{\tiny $\pm 2.9$}
\\
InfoGraph &
29.3{\tiny $\pm 3.0$}
&
28.0{\tiny $\pm 0.7$}
&
26.0{\tiny $\pm 2.3$}
&
18.0{\tiny $\pm 3.5$}
&
22.7{\tiny $\pm 1.4$}
&
26.7{\tiny $\pm 2.3$}
&
28.7{\tiny $\pm 3.5$}
&
25.3{\tiny $\pm 3.2$}
&
24.0{\tiny $\pm 2.8$}
&
28.7{\tiny $\pm 2.9$}
&
26.0{\tiny $\pm 2.8$}
&
28.7{\tiny $\pm 4.2$}
&
26.0{\tiny $\pm 2.7$}
\\
TGNN &
27.3{\tiny $\pm 1.9$}
&
26.7{\tiny $\pm 2.7$}
&
27.3{\tiny $\pm 3.8$}
&
20.0{\tiny $\pm 1.1$}
&
18.7{\tiny $\pm 2.5$}
&
14.7{\tiny $\pm 2.5$}
&
27.3{\tiny $\pm 3.6$}
&
25.3{\tiny $\pm 1.0$}
&
30.0{\tiny $\pm 2.0$}
&
28.7{\tiny $\pm 2.7$}
&
24.7{\tiny $\pm 0.6$}
&
26.7{\tiny $\pm 1.0$}
&
24.8{\tiny $\pm 2.1$}
\\
PLUE &
29.8{\tiny $\pm 2.3$}
&
24.5{\tiny $\pm 2.2$}
&
26.2{\tiny $\pm 3.0$}
&
21.5{\tiny $\pm 2.1$}
&
20.7{\tiny $\pm 2.4$}
&
22.7{\tiny $\pm 2.7$}
&
29.0{\tiny $\pm 2.0$}
&
24.3{\tiny $\pm 1.3$}
&
27.0{\tiny $\pm 3.1$}
&
30.3{\tiny $\pm 3.0$}
&
25.3{\tiny $\pm 2.3$}
&
31.5{\tiny $\pm 2.7$}
&
26.1{\tiny $\pm 2.4$}
\\
\midrule 
\rowcolor{LightCyan}  \textbf{\method{}} &
33.3{\tiny $\pm 2.6$}
&
26.7{\tiny $\pm 2.3$}
&
27.3{\tiny $\pm 2.6$}
&
26.7{\tiny $\pm 1.8$}
&
21.3{\tiny $\pm 1.5$}
&
23.3{\tiny $\pm 1.7$}
&
36.7{\tiny $\pm 2.0$}
&
24.7{\tiny $\pm 2.9$}
&
34.7{\tiny $\pm 2.9$}
&
35.3{\tiny $\pm 2.3$}
&
34.0{\tiny $\pm 1.3$}
&
44.7{\tiny $\pm 2.5$}
&
\textbf{30.7}{\tiny $\pm 2.2$}
\\
\bottomrule[1pt]
\end{tabular}}
\label{tab::1}
\end{table*}

\begin{table*}[t]
\centering
\tabcolsep=2pt
\caption{The classification results (in \%) on Mutagenicity (source$\rightarrow$target). M0, M1, M2, and M3 are split by the graph density.}
\resizebox{\textwidth}{!}{
\begin{tabular}{llllllllllllll}
\toprule[1pt]
{\bf Methods} &M0$\rightarrow$M1 &M1$\rightarrow$M0 &M0$\rightarrow$M2 &M2$\rightarrow$M0 &M0$\rightarrow$M3 &M3$\rightarrow$M0 &M1$\rightarrow$M2 &M2$\rightarrow$M1 &M1$\rightarrow$M3 &M3$\rightarrow$M1 &M2$\rightarrow$M3 &M3$\rightarrow$M2 &Avg.\\
\midrule
\midrule
GCN &
68.0{\tiny $\pm 2.0$}
&
68.8{\tiny $\pm 1.5$}
&
60.5{\tiny $\pm 2.8$}
&
64.4{\tiny $\pm 1.5$}
&
53.7{\tiny $\pm 1.3$}
&
58.1{\tiny $\pm 1.4$}
&
75.2{\tiny $\pm 0.8$}
&
76.2{\tiny $\pm 1.5$}
&
67.5{\tiny $\pm 1.2$}
&
55.4{\tiny $\pm 1.5$}
&
62.5{\tiny $\pm 1.0$}
&
68.5{\tiny $\pm 1.5$}
&
64.9{\tiny $\pm 1.5$}
\\
GIN &
70.6{\tiny $\pm 0.4$}
&
64.2{\tiny $\pm 0.9$}
&
63.5{\tiny $\pm 1.2$}
&
62.5{\tiny $\pm 0.7$}
&
57.0{\tiny $\pm 0.1$}
&
56.4{\tiny $\pm 0.3$}
&
73.3{\tiny $\pm 0.5$}
&
76.5{\tiny $\pm 0.4$}
&
65.2{\tiny $\pm 0.6$}
&
53.3{\tiny $\pm 1.8$}
&
64.4{\tiny $\pm 0.8$}
&
66.8{\tiny $\pm 0.5$}
&
64.5{\tiny $\pm 0.7$}
\\
GraphSAGE &
71.2{\tiny $\pm 0.6$}
&
65.6{\tiny $\pm 0.3$}
&
64.3{\tiny $\pm 0.3$}
&
65.5{\tiny $\pm 0.2$}
&
57.3{\tiny $\pm 0.5$}
&
56.5{\tiny $\pm 0.3$}
&
74.7{\tiny $\pm 0.4$}
&
77.6{\tiny $\pm 0.7$}
&
62.3{\tiny $\pm 0.4$}
&
51.7{\tiny $\pm 0.3$}
&
62.4{\tiny $\pm 0.5$}
&
62.3{\tiny $\pm 0.7$}
&
64.3{\tiny $\pm 0.4$}
\\
GAT &
69.7{\tiny $\pm 1.0$}
&
67.0{\tiny $\pm 1.6$}
&
62.7{\tiny $\pm 2.3$}
&
67.0{\tiny $\pm 1.5$}
&
56.1{\tiny $\pm 2.1$}
&
57.8{\tiny $\pm 1.7$}
&
76.6{\tiny $\pm 1.2$}
&
77.2{\tiny $\pm 0.4$}
&
63.4{\tiny $\pm 1.3$}
&
53.0{\tiny $\pm 3.9$}
&
60.7{\tiny $\pm 0.6$}
&
61.8{\tiny $\pm 3.2$}
&
64.6{\tiny $\pm 1.7$}
\\
Mean-Teacher &
65.3{\tiny $\pm 4.7$}
&
52.1{\tiny $\pm 3.4$}
&
70.6{\tiny $\pm 2.5$}
&
52.2{\tiny $\pm 1.9$}
&
49.9{\tiny $\pm 0.5$}
&
49.0{\tiny $\pm 0.4$}
&
66.2{\tiny $\pm 1.4$}
&
62.7{\tiny $\pm 4.1$}
&
50.1{\tiny $\pm 1.3$}
&
72.2{\tiny $\pm 1.6$}
&
48.9{\tiny $\pm 1.7$}
&
65.8{\tiny $\pm 3.1$}
&
58.7{\tiny $\pm 2.2$}
\\
InfoGraph &
69.1{\tiny $\pm 1.8$}
&
68.9{\tiny $\pm 0.3$}
&
66.6{\tiny $\pm 2.5$}
&
64.9{\tiny $\pm 1.2$}
&
55.9{\tiny $\pm 1.0$}
&
57.8{\tiny $\pm 2.1$}
&
74.7{\tiny $\pm 0.3$}
&
76.8{\tiny $\pm 1.5$}
&
65.6{\tiny $\pm 0.6$}
&
57.1{\tiny $\pm 3.2$}
&
64.7{\tiny $\pm 1.9$}
&
64.2{\tiny $\pm 2.9$}
&
65.5{\tiny $\pm 1.6$}
\\
TGNN &
73.3{\tiny $\pm 4.9$}
&
61.9{\tiny $\pm 2.4$}
&
65.3{\tiny $\pm 4.0$}
&
58.1{\tiny $\pm 2.4$}
&
55.5{\tiny $\pm 3.5$}
&
58.1{\tiny $\pm 2.4$}
&
65.9{\tiny $\pm 1.1$}
&
66.7{\tiny $\pm 3.9$}
&
66.5{\tiny $\pm 2.1$}
&
70.1{\tiny $\pm 1.0$}
&
55.5{\tiny $\pm 3.5$}
&
65.3{\tiny $\pm 3.0$}
&
63.5{\tiny $\pm 2.9$}
\\
PLUE &
75.2{\tiny $\pm 1.4$}
&
68.5{\tiny $\pm 0.5$}
&
66.3{\tiny $\pm 1.1$}
&
67.9{\tiny $\pm 1.6$}
&
54.0{\tiny $\pm 1.3$}
&
56.4{\tiny $\pm 1.4$}
&
68.4{\tiny $\pm 1.0$}
&
76.9{\tiny $\pm 3.5$}
&
62.9{\tiny $\pm 0.4$}
&
57.6{\tiny $\pm 3.0$}
&
62.0{\tiny $\pm 0.6$}
&
67.4{\tiny $\pm 2.6$}
&
65.3{\tiny $\pm 1.5$}
\\
\midrule
\rowcolor{LightCyan} \textbf{\method{}} &
76.4{\tiny $\pm 0.8$}
&
69.6{\tiny $\pm 1.3$}
&
70.0{\tiny $\pm 2.2$}
&
63.2{\tiny $\pm 1.2$}
&
58.4{\tiny $\pm 1.2$}
&
60.6{\tiny $\pm 1.3$}
&
76.9{\tiny $\pm 1.8$}
&
80.1{\tiny $\pm 1.9$}
&
65.7{\tiny $\pm 1.1$}
&
66.5{\tiny $\pm 3.1$}
&
65.6{\tiny $\pm 1.5$}
&
70.6{\tiny $\pm 1.3$}
&
\textbf{68.6}{\tiny $\pm 1.6$}
\\
\bottomrule[1pt]
\end{tabular}
}
\label{tab::2}
\end{table*}

\begin{figure*}[!ht]
    \centering
    \includegraphics[width=\textwidth]{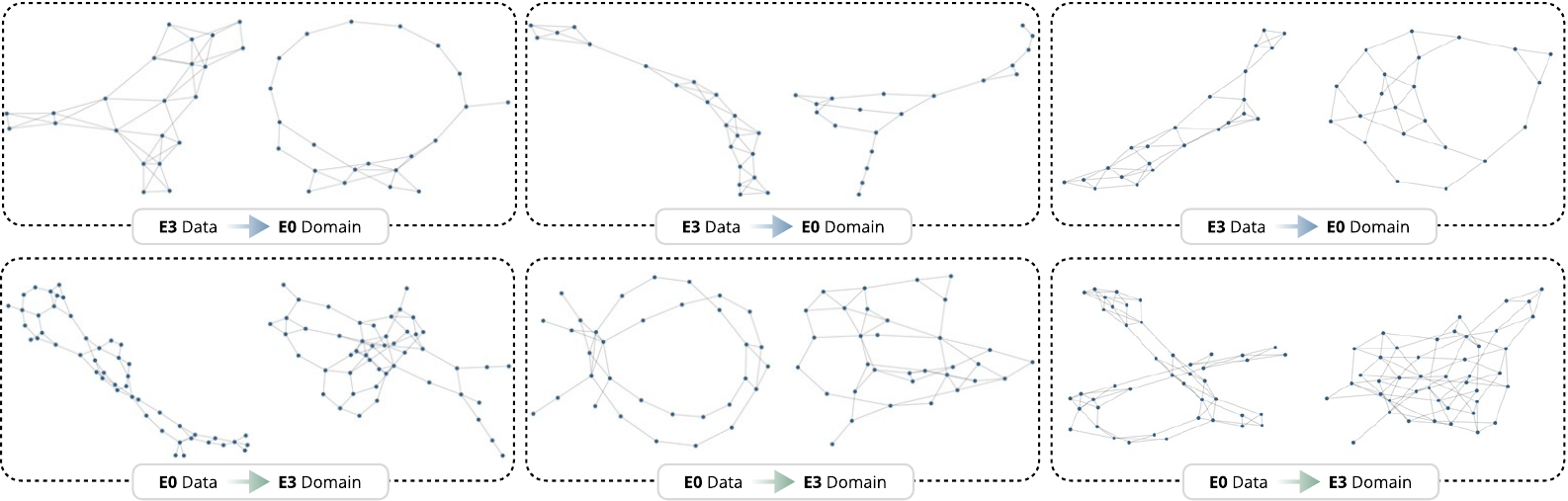}
    \caption{Visualization of diffusion adaptation on ENZYMES.
    E0 and E3 are subsets of ENZYMES, with E0 being sparser and E3 being denser. 
    The diffusion process can reconstruct the graph while preserving semantics. 
    As $\mathbf{E3\rightarrow E0}$, the graphs become sparser. 
    As $\mathbf{E0\rightarrow E3}$, the graphs become denser. }
    \label{fig:sup-diff-enzymes}
\end{figure*}

\begin{table*}[tb]
\centering
\tabcolsep=2pt
\caption{The classification results (in \%) on PROTEINS (source$\rightarrow$target). P0, P1, P2, and P3 are split by the graph density.}
\resizebox{\textwidth}{!}{
\begin{tabular}{llllllllllllll}
\toprule[1pt]
{\bf Methods} &P0$\rightarrow$P1 &P1$\rightarrow$P0 &P0$\rightarrow$P2 &P2$\rightarrow$P0 &P0$\rightarrow$P3 &P3$\rightarrow$P0 &P1$\rightarrow$P2 &P2$\rightarrow$P1 &P1$\rightarrow$P3 &P3$\rightarrow$P1 &P2$\rightarrow$P3 &P3$\rightarrow$P2 &Avg.\\
\midrule
\midrule
GCN &
71.9{\tiny $\pm 0.9$}
&
74.7{\tiny $\pm 2.9$}
&
62.6{\tiny $\pm 1.2$}
&
68.3{\tiny $\pm 3.8$}
&
51.1{\tiny $\pm 3.3$}
&
45.8{\tiny $\pm 3.1$}
&
57.6{\tiny $\pm 2.1$}
&
70.4{\tiny $\pm 2.2$}
&
39.7{\tiny $\pm 3.6$}
&
49.7{\tiny $\pm 2.6$}
&
58.3{\tiny $\pm 1.3$}
&
52.9{\tiny $\pm 3.0$}
&
58.6{\tiny $\pm 2.5$}
\\
GIN &
70.0{\tiny $\pm 2.1$}
&
60.7{\tiny $\pm 3.6$}
&
61.8{\tiny $\pm 2.6$}
&
72.9{\tiny $\pm 2.7$}
&
47.1{\tiny $\pm 3.3$}
&
44.3{\tiny $\pm 4.2$}
&
62.5{\tiny $\pm 2.1$}
&
68.9{\tiny $\pm 2.0$}
&
41.1{\tiny $\pm 3.2$}
&
47.9{\tiny $\pm 3.3$}
&
48.6{\tiny $\pm 4.0$}
&
56.1{\tiny $\pm 2.6$}
&
57.4{\tiny $\pm 3.0$}
\\
GraphSAGE &
72.2{\tiny $\pm 1.0$}
&
78.3{\tiny $\pm 3.0$}
&
64.7{\tiny $\pm 2.3$}
&
67.2{\tiny $\pm 1.1$}
&
46.9{\tiny $\pm 1.1$}
&
42.2{\tiny $\pm 4.0$}
&
62.6{\tiny $\pm 1.8$}
&
69.7{\tiny $\pm 0.8$}
&
32.9{\tiny $\pm 2.0$}
&
50.8{\tiny $\pm 2.3$}
&
56.1{\tiny $\pm 3.5$}
&
56.9{\tiny $\pm 3.4$}
&
58.9{\tiny $\pm 2.2$}
\\
GAT &
70.0{\tiny $\pm 3.7$}
&
71.4{\tiny $\pm 3.7$}
&
66.8{\tiny $\pm 2.0$}
&
73.9{\tiny $\pm 2.6$}
&
49.3{\tiny $\pm 1.3$}
&
40.4{\tiny $\pm 2.8$}
&
61.4{\tiny $\pm 4.6$}
&
68.9{\tiny $\pm 2.2$}
&
44.3{\tiny $\pm 3.8$}
&
49.3{\tiny $\pm 3.6$}
&
50.7{\tiny $\pm 2.4$}
&
52.1{\tiny $\pm 2.9$}
&
58.3{\tiny $\pm 3.0$}
\\
Mean-Teacher &
64.3{\tiny $\pm 4.1$}
&
71.4{\tiny $\pm 5.2$}
&
60.4{\tiny $\pm 3.6$}
&
72.1{\tiny $\pm 4.4$}
&
25.0{\tiny $\pm 5.6$}
&
55.4{\tiny $\pm 4.0$}
&
61.1{\tiny $\pm 2.3$}
&
60.7{\tiny $\pm 5.3$}
&
29.6{\tiny $\pm 4.8$}
&
49.3{\tiny $\pm 3.1$}
&
31.8{\tiny $\pm 4.0$}
&
55.4{\tiny $\pm 4.9$}
&
51.0{\tiny $\pm 4.3$}
\\
InfoGraph &
74.0{\tiny $\pm 2.7$}
&
77.6{\tiny $\pm 2.9$}
&
68.3{\tiny $\pm 3.6$}
&
71.1{\tiny $\pm 1.1$}
&
46.9{\tiny $\pm 3.2$}
&
46.5{\tiny $\pm 2.1$}
&
64.4{\tiny $\pm 1.5$}
&
72.2{\tiny $\pm 1.9$}
&
41.9{\tiny $\pm 1.1$}
&
35.4{\tiny $\pm 1.9$}
&
54.7{\tiny $\pm 3.1$}
&
62.6{\tiny $\pm 4.2$}
&
59.4{\tiny $\pm 2.4$}
\\
TGNN &
60.4{\tiny $\pm 3.4$}
&
41.8{\tiny $\pm 2.8$}
&
45.7{\tiny $\pm 4.6$}
&
46.6{\tiny $\pm 4.2$}
&
64.8{\tiny $\pm 5.4$}
&
42.0{\tiny $\pm 3.9$}
&
58.6{\tiny $\pm 3.2$}
&
58.4{\tiny $\pm 3.7$}
&
75.7{\tiny $\pm 4.0$}
&
68.4{\tiny $\pm 3.2$}
&
65.5{\tiny $\pm 4.8$}
&
53.8{\tiny $\pm 2.0$}
&
55.9{\tiny $\pm 3.8$}
\\
PLUE &
64.3{\tiny $\pm 2.3$}
&
71.5{\tiny $\pm 2.1$}
&
65.0{\tiny $\pm 3.0$}
&
78.4{\tiny $\pm 2.1$}
&
45.9{\tiny $\pm 3.7$}
&
63.7{\tiny $\pm 3.0$}
&
58.3{\tiny $\pm 3.3$}
&
68.9{\tiny $\pm 2.7$}
&
41.9{\tiny $\pm 1.3$}
&
45.7{\tiny $\pm 4.1$}
&
47.5{\tiny $\pm 3.2$}
&
58.3{\tiny $\pm 2.7$}
&
59.1{\tiny $\pm 2.8$}
\\
\midrule 
\rowcolor{LightCyan} \textbf{\method{}} &
72.3{\tiny $\pm 1.0$}
&
73.9{\tiny $\pm 2.0$}
&
66.8{\tiny $\pm 3.4$}
&
78.5{\tiny $\pm 1.9$}
&
47.1{\tiny $\pm 2.2$}
&
65.6{\tiny $\pm 3.7$}
&
61.1{\tiny $\pm 1.1$}
&
71.8{\tiny $\pm 3.1$}
&
43.9{\tiny $\pm 2.0$}
&
50.4{\tiny $\pm 2.6$}
&
43.2{\tiny $\pm 2.0$}
&
60.0{\tiny $\pm 3.4$}
&
\textbf{61.2}{\tiny $\pm 2.4$}
\\
\bottomrule[1pt]
\end{tabular}}
\label{tab::3}
\end{table*}

\begin{table*}[t]
\centering
\tabcolsep=2pt
\caption{The classification results (in \%) on FRANKENSTEIN (source$\rightarrow$target). F0, F1, F2, and F3 are split by the graph density.}
\resizebox{\textwidth}{!}{
\begin{tabular}{llllllllllllll}
\toprule[1pt]
{\bf Methods} &F0$\rightarrow$F1 &F1$\rightarrow$F0 &F0$\rightarrow$F2 &F2$\rightarrow$F0 &F0$\rightarrow$F3 &F3$\rightarrow$F0 &F1$\rightarrow$F2 &F2$\rightarrow$F1 &F1$\rightarrow$F3 &F3$\rightarrow$F1 &F2$\rightarrow$F3 &F3$\rightarrow$F2 &Avg.\\
\midrule
\midrule
GCN &
55.3{\tiny $\pm 0.8$} 
&
56.4{\tiny $\pm 1.6$} 
&
60.4{\tiny $\pm 1.9$}
&
54.6{\tiny $\pm 1.4$}
&
46.7{\tiny $\pm 1.8$} 
&
51.6{\tiny $\pm 1.7$}
&
60.7{\tiny $\pm 0.8$}
&
58.3{\tiny $\pm 1.3$} 
&
47.9{\tiny $\pm 1.8$}
&
47.5{\tiny $\pm 0.8$}
&
52.1{\tiny $\pm 2.6$} 
&
54.6{\tiny $\pm 1.8$} 
&
53.8{\tiny $\pm 1.5$} 
\\
GIN &
56.5{\tiny $\pm 0.5$}
&
53.8{\tiny $\pm 1.4$}
&
57.2{\tiny $\pm 0.4$}
&
57.9{\tiny $\pm 1.8$}
&
50.7{\tiny $\pm 3.5$}
&
51.0{\tiny $\pm 0.6$}
&
58.7{\tiny $\pm 1.3$}
&
58.3{\tiny $\pm 0.8$}
&
46.7{\tiny $\pm 1.4$}
&
47.9{\tiny $\pm 1.3$}
&
50.5{\tiny $\pm 0.7$}
&
52.0{\tiny $\pm 2.1$}
&
53.4{\tiny $\pm 1.3$} 
\\
GraphSAGE &
56.6{\tiny $\pm 1.3$}
&
53.5{\tiny $\pm 0.4$}
&
55.4{\tiny $\pm 1.0$}
&
57.8{\tiny $\pm 1.3$}
&
49.9{\tiny $\pm 0.3$}
&
55.5{\tiny $\pm 1.6$}
&
59.4{\tiny $\pm 1.4$}
&
59.6{\tiny $\pm 0.2$}
&
47.1{\tiny $\pm 1.0$}
&
49.3{\tiny $\pm 0.8$}
&
49.2{\tiny $\pm 1.7$}
&
52.7{\tiny $\pm 1.3$}
&
53.8{\tiny $\pm 1.0$} 
\\
GAT &
57.1{\tiny $\pm 0.9$}
&
56.0{\tiny $\pm 0.8$}
&
58.5{\tiny $\pm 1.8$}
&
56.4{\tiny $\pm 2.2$}
&
48.7{\tiny $\pm 1.1$}
&
51.6{\tiny $\pm 2.8$}
&
62.5{\tiny $\pm 1.8$}
&
57.1{\tiny $\pm 0.7$}
&
46.2{\tiny $\pm 2.2$}
&
47.2{\tiny $\pm 1.4$}
&
51.7{\tiny $\pm 2.6$}
&
53.6{\tiny $\pm 0.9$}
&
53.9{\tiny $\pm 1.6$} 
\\
Mean-Teacher &
57.0{\tiny $\pm 4.6$}
&
53.8{\tiny $\pm 3.3$}
&
55.6{\tiny $\pm 3.5$}
&
54.2{\tiny $\pm 2.5$}
&
47.6{\tiny $\pm 3.8$}
&
49.7{\tiny $\pm 1.5$}
&
56.2{\tiny $\pm 3.4$}
&
59.1{\tiny $\pm 4.8$}
&
48.5{\tiny $\pm 3.4$}
&
52.9{\tiny $\pm 5.2$}
&
51.4{\tiny $\pm 3.3$}
&
53.1{\tiny $\pm 4.7$}
&
53.2{\tiny $\pm 3.7$} 
\\
InfoGraph &
57.0{\tiny $\pm 2.7$}
&
55.7{\tiny $\pm 2.2$}
&
60.1{\tiny $\pm 2.6$}
&
60.0{\tiny $\pm 3.0$}
&
48.9{\tiny $\pm 2.0$}
&
51.2{\tiny $\pm 1.7$}
&
60.6{\tiny $\pm 1.1$}
&
61.8{\tiny $\pm 1.9$}
&
45.4{\tiny $\pm 2.3$}
&
46.3{\tiny $\pm 1.2$}
&
53.2{\tiny $\pm 2.0$}
&
53.5{\tiny $\pm 0.8$}
&
54.5{\tiny $\pm 1.9$} 
\\
TGNN &
53.2{\tiny $\pm 5.6$}
&
48.8{\tiny $\pm 1.7$}
&
54.0{\tiny $\pm 5.2$}
&
54.2{\tiny $\pm 1.7$}
&
46.2{\tiny $\pm 4.0$}
&
47.9{\tiny $\pm 0.5$}
&
55.1{\tiny $\pm 5.1$}
&
50.8{\tiny $\pm 3.6$}
&
56.8{\tiny $\pm 4.0$}
&
53.2{\tiny $\pm 5.6$}
&
56.8{\tiny $\pm 3.0$}
&
48.1{\tiny $\pm 4.5$}
&
53.1{\tiny $\pm 3.7$} 
\\
PLUE &
57.9{\tiny $\pm 1.2$}
&
56.4{\tiny $\pm 1.3$}
&
60.0{\tiny $\pm 1.9$}
&
59.1{\tiny $\pm 1.6$}
&
49.1{\tiny $\pm 0.6$}
&
53.2{\tiny $\pm 1.8$}
&
60.8{\tiny $\pm 1.5$}
&
52.3{\tiny $\pm 3.7$}
&
48.1{\tiny $\pm 3.7$}
&
52.1{\tiny $\pm 4.1$}
&
52.7{\tiny $\pm 1.5$}
&
53.9{\tiny $\pm 2.2$}
&
54.6{\tiny $\pm 2.1$} 
\\
\midrule
\rowcolor{LightCyan}
{\bf \method{}} &
59.7{\tiny $\pm 0.6$}
&
56.9{\tiny $\pm 0.9$}
&
58.8{\tiny $\pm 1.1$}
&
53.8{\tiny $\pm 0.4$}
&
51.7{\tiny $\pm 1.2$}
&
55.9{\tiny $\pm 1.7$}
&
61.0{\tiny $\pm 0.5$}
&
62.7{\tiny $\pm 0.7$}
&
53.0{\tiny $\pm 0.6$}
&
54.5{\tiny $\pm 1.0$}
&
53.5{\tiny $\pm 0.5$}
&
56.4{\tiny $\pm 0.8$}
&
\textbf{56.5}{\tiny $\pm 0.8$}
\\
\bottomrule[1pt]
\end{tabular}}
\label{tab::4}
\end{table*}

\begin{figure*}[t]
    \centering
    \includegraphics[width=\textwidth]{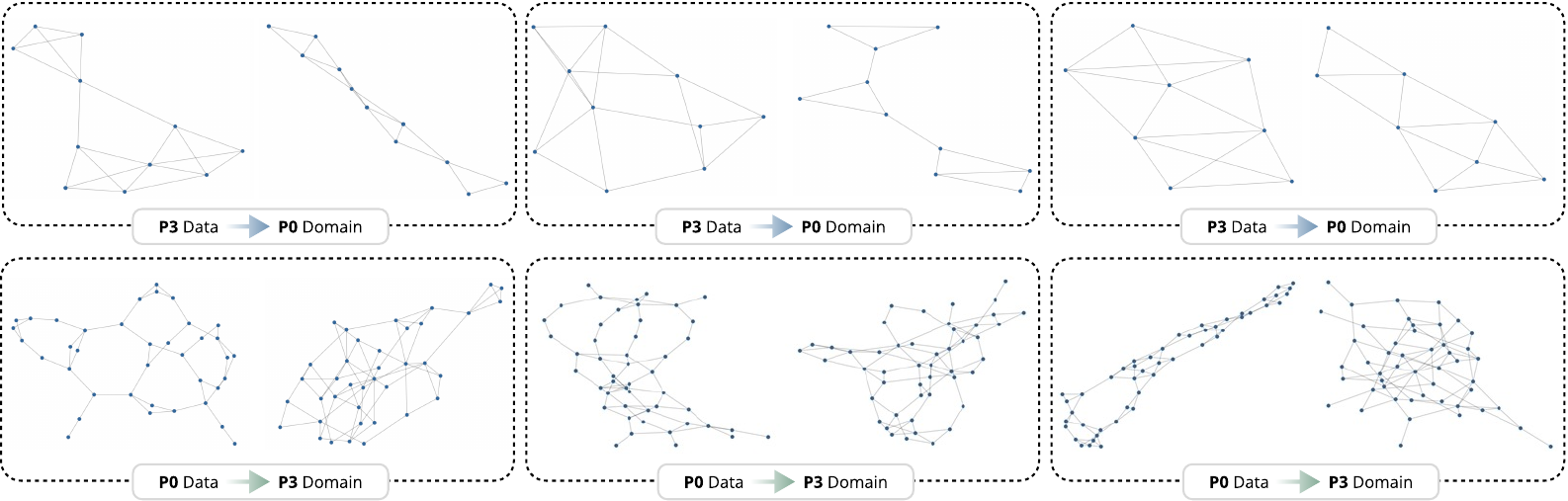}
    \caption{Visualization of diffusion adaptation on PROTEINS.
    P0 and P3 are subsets of PROTEINS, with P0 being sparser and P3 being denser. 
    The diffusion process can reconstruct the graph while preserving semantics. 
    As $\mathbf{P3\rightarrow P0}$, the graphs become sparser. 
    As $\mathbf{P0\rightarrow P3}$, the graphs become denser. }
    \label{fig:sup-diff-proteins}
\end{figure*}

\begin{figure*}[ht]
    \centering
    \includegraphics[width=0.9\textwidth]{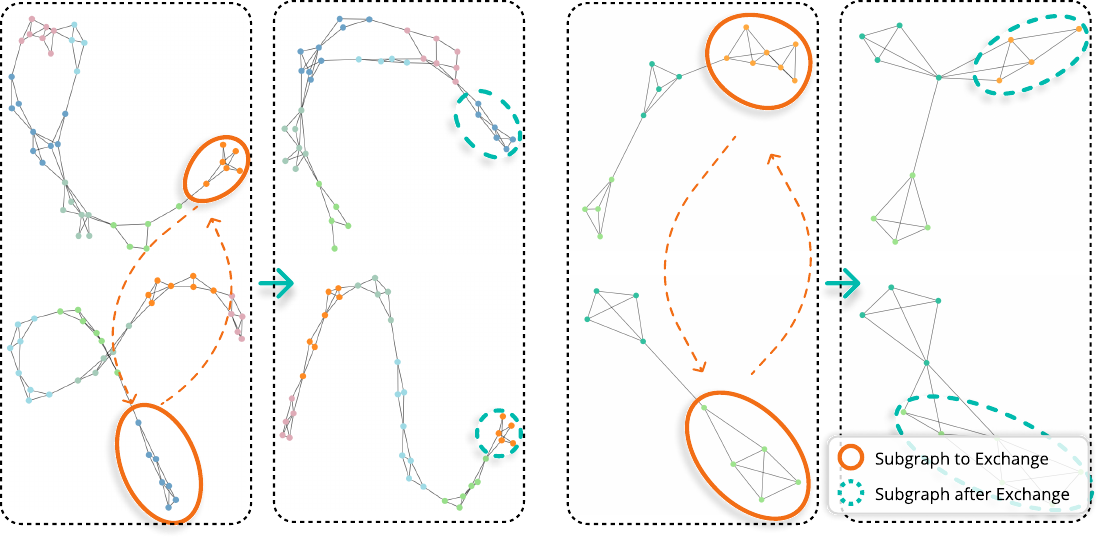}
    \caption{Visualization of graph jigsaw on ENZYMES. Graph jigsaw, combined with consistency learning, leverages unlabeled target data to enhance the generalization capacity.}
    \label{fig:sup-vis-jigsaw}
\end{figure*}

\begin{table}[!ht]
\centering
\tabcolsep=2pt
\caption{The classification results (in \%) on COX2 and BZR~(source$\rightarrow$target). C, CM, B, and BM are for COX2, COX2\_MD, BZR, and BZR\_MD datasets.}
\resizebox{\linewidth}{!}{
\begin{tabular}{llllll}
\toprule[1pt]
{\bf Methods} &C$\rightarrow$CM & CM$\rightarrow$C &B$\rightarrow$BM &BM$\rightarrow$B &Avg.\\
\midrule
\midrule
GCN &
54.1{\tiny $\pm 2.6$}&
46.6 {\tiny $\pm 4.1$}&
51.3 {\tiny $\pm 2.3$}&
62.8 {\tiny $\pm 3.7$}&
53.7{\tiny $\pm 3.2$}
\\
GIN &
51.1{\tiny $\pm 2.2$}&
46.4{\tiny $\pm 4.6$}&
48.1{\tiny $\pm 3.6$}&
65.6{\tiny $\pm 2.8$}&
52.8{\tiny $\pm 3.3$}
\\
GraphSAGE &
49.2{\tiny $\pm 3.4$}&
42.9{\tiny $\pm 3.9$}&
47.3{\tiny $\pm 1.5$}&
67.5{\tiny $\pm 3.1$}&
51.7{\tiny $\pm 3.0$}
\\
GAT &
52.0{\tiny $\pm 1.8$}&
48.9{\tiny $\pm 3.7$}&
48.4{\tiny $\pm 2.2$}&
61.3{\tiny $\pm 4.2$}&
52.6{\tiny $\pm 3.0$}
\\
Mean-Teacher &
53.0{\tiny $\pm 2.3$}&
42.6{\tiny $\pm 4.9$}&
50.6{\tiny $\pm 2.1$}&
57.6{\tiny $\pm 4.3$}&
50.9{\tiny $\pm 3.4$}
\\
InfoGraph &
45.9{\tiny $\pm 3.4$}&
48.9{\tiny $\pm 3.3$}&
51.9{\tiny $\pm 3.2$}&
65.2{\tiny $\pm 4.7$}&
53.4{\tiny $\pm 3.6$}
\\
TGNN &
48.3{\tiny $\pm 4.2$}&
52.1{\tiny $\pm 5.6$}&
46.4{\tiny $\pm 2.7$}&
68.5{\tiny $\pm 4.3$}&
53.8{\tiny $\pm 4.2$}
\\
PLUE &
54.4{\tiny $\pm 1.4$}&
41.7{\tiny $\pm 3.2$}&
49.8{\tiny $\pm 2.8$}&
74.8{\tiny $\pm 1.5$}&
55.2{\tiny $\pm 2.2$}
\\
\midrule
\rowcolor{LightCyan}
{\bf \method{}} &
56.6{\tiny $\pm 0.7$}&
59.1{\tiny $\pm 2.6$}&
53.2{\tiny $\pm 2.0$}&
73.2{\tiny $\pm 3.1$}&
\textbf{60.5}{\tiny $\pm 2.1$}
\\
\bottomrule[1pt]
\end{tabular}
}
\label{tab::cross}
\end{table}

\begin{table*}[!ht]
\centering
\caption{Ablation studies of different submodules.}
\begin{tabular}{c|c|ccccccc}
\toprule[1pt]
& Methods & ENZYMES & Mutagenicity & PROTEINS & FRANKENSTEIN  & COX2 & BZR & Avg. \\ 
\midrule
\midrule
$M_1$&\method{} \textit{w/o} GDM & 27.7{\tiny $\pm 3.1$} & 67.2{\tiny $\pm 1.8$} & 59.7{\tiny $\pm 3.7$} & 55.6{\tiny $\pm 0.7$}  & 56.1{\tiny $\pm 1.9$} & 62.0{\tiny $\pm 3.4$} & 54.7{\tiny $\pm 2.4$} \\
$M_2$&\method{} \textit{w/o} UPCL & 28.3{\tiny $\pm 1.5$} & 67.5{\tiny $\pm 1.5$} & 60.2{\tiny $\pm 2.1$} & 55.4{\tiny $\pm 0.5$} & 56.5{\tiny $\pm 1.6$} & 62.3{\tiny $\pm 2.9$} & 55.0{\tiny $\pm 1.7$}\\
$M_3$&\method{} \textit{w/o} CLGJ & 28.7{\tiny $\pm 2.5$} & 67.9{\tiny $\pm 2.0$} & 60.0{\tiny $\pm 2.4$} & 56.1{\tiny $\pm 0.9$} & 57.2{\tiny $\pm 2.2$} & 63.0{\tiny $\pm 2.8$} & 55.5{\tiny $\pm 2.1$}\\ 
\midrule
\rowcolor{LightCyan} $M_4$&\method{}~(ours) & 30.7{\tiny $\pm 2.2$} & 68.6{\tiny $\pm 1.6$} & 61.2{\tiny $\pm 2.4$} & 56.5{\tiny $\pm 0.8$} & 57.9{\tiny $\pm 1.7$} & 63.2{\tiny $\pm 2.6$} & 56.3{\tiny $\pm 1.9$} \\ \bottomrule[1pt]
\end{tabular}
\label{tab::abl}
\end{table*}

\subsection{Experimental Settings}

\paratitle{Datasets.}
We perform experiments on real-world benchmark datasets with source-free domain adaption settings. In order to demonstrate the performance of our approach in various scenarios, we conducted experiments on both dataset split and cross-dataset source free domain adaptation. In ENZYMES, Mutagenicity, PROTEINS, and FRANKENSTEIN datasets, we split the data, thereby introducing domain discrepancies. Then, we perform source-free domain adaptation across the sub-datasets. As for the COX2 and BZR datasets, we directly conduct source-free domain adaptation on the sub-datasets associated with each of them.
\begin{itemize}
    \item \textbf{ENZYMES}~\cite{enzymes} is a bioinformatic data set comprising 600 tertiary protein structures. It constitutes a classic collection of data rooted in the structural information of biological molecules and proteins.
    \item \textbf{Mutagenicity}~\cite{mutagenicity}  encompasses a multitude of molecular structures, each intricately paired with its corresponding Ames test data, amounting to a total of 4337 molecular structures. 
    \item \textbf{PROTEINS}~\cite{proteins} contain protein data in graph form, where each label indicates whether a protein is a non-enzyme. The amino acids serve as nodes. The edges exist when the distance between two nodes is less than 6 angstroms. 
    \item \textbf{FRANKENSTEIN}~\cite{orsini2015graph} is a composite dataset that amalgamates the BURSI and MNIST datasets. Each data sample is depicted as a graph, with the chemical atom symbols representing vertices and the bond types representing edges.
    \item \textbf{COX2}~\cite{sutherland2003spline}. We engage the COX2 and COX2\_MD datasets, consisting of 467 and 303 inhibitors specifically targeting cyclooxygenase-2. Within these datasets, individual graphs serve as graphical representations of distinct chemical compounds. These graphs feature edges annotated with distance values, while the vertex labels are indicative of the atom types found in the compounds.
    \item \textbf{BZR}~\cite{sutherland2003spline}. We use BZR and BZR\_MD datasets, which consist of 405 and 306 ligands designed to interact with the benzodiazepine receptor. The process of graph construction employed in these datasets is analogous to that of the COX2 dataset.
\end{itemize}

\paratitle{Domain Settings}. This work follows the setting of source-free domain adaptation~\cite{saito2017asymmetric, shot} in the image classification task. The division of the dataset is executed by using graph density, which is defined as the ratio of the number of existing edges to the number of potential edges.
It can be calculated as 
\begin{equation}
    D = \frac{2|E|}{|V|(|V|-1)},
\end{equation}
where $|E|$ is the number of edges and $|V|$ is the number of nodes in the graph $G$.
Within each sub-dataset, the ratio of the training set to the test set is 8:2.
Each dataset is split according to the above instructions, leading to domain discrepancies among these sub-datasets and subsequently yield inferior cross-domain performance. 

Diffusion models and pre-trained GNNs are trained on the source domain as off-the-shelf models to simulate real-world scenarios. The off-the-shelf models adapt to the target domain. In the adaptation process, only the target domain data is used, and the source domain data is not available.  This is more aligned with real-world application scenarios.

\paratitle{Baselines.}
For credibility, we compare \method{} with extensive competing baselines, which can be categorized into three aspects: (a) Graph Neural Networks, including GCN~\cite{GCN}, GIN~\cite{GIN}, GAT~\cite{GAT} and GraphSAGE~\cite{GraphSAGE}. (b) Graph Semi-Supervised Methods, including Mean-Teacher~\cite{tarvainen2017mean}, InfoGraph~\cite{sun2020infograph} and TGNN~\cite{tgnn}. (c) Source-free Adaptation Methods. We adopt a recent state-of-the-art source-free domain adoption method PLUE~\cite{plue} of image classification.
Note that the semi-supervised methods are designed to utilize both source and target data. Therefore, the comparison with semi-supervised methods may not be entirely fair.
\begin{itemize}
    \item \textbf{GCN}~\cite{GCN} adapts localized first-order approximation of spectral graph convolution to represent the graph data.
    \item \textbf{GIN}~\cite{GIN} takes the message passing neural networks for capturing the different topological structures of graphs.
    \item \textbf{GraphSAGE}~\cite{GraphSAGE} leverages node feature information to efficiently generate node embeddings for unseen data.
    \item \textbf{GAT}~\cite{GAT} introduces the attention mechanism, to facilitate the model focusing on the most informative parts.
    \item \textbf{Mean-Teacher}~\cite{tarvainen2017mean} is an effective semi-supervised algorithm in the graph domain. The method uses a student to make predictions and a teacher model to generate training targets. The teacher model is an exponential moving average of the student model.
    \item \textbf{InfoGraph}~\cite{sun2020infograph} is a semi-supervised method for GNNs that uses mutual information maximization to learn representations of graphs.
    \item \textbf{TGNN}~(Twin Graph Neural Network)~\cite{tgnn} is a semi-supervised approach for graph classification, which contains a message-passing module and a graph kernel module to explore structural information.
    \item \textbf{PLUE}~\cite{plue} is a recent state-of-the-art source-free domain adaption method for the image classification task, which proposes to evaluate the reliability of refined labels while excluding the noisy label. We implement PLUE in graph classification tasks. 
\end{itemize}

\subsection{Implementation Details}\label{sec:imple}

In this passage, we introduce the implementation details of \method{} at different stages. The baseline methods are initiated with hyperparameters as the corresponding paper and fine-tuned to achieve the best performance. To mitigate the potential impact of randomness, we employ $5$ runs and report the average accuracy and standard deviation.

\paratitle{Model Training on Souce Domain.}
We adopt GCN as the default GNN encoder, with the embedding dimension $64$ and the layer number $3$. We use Adam optimizer and a learning rate of $0.001$. We train the model in the source data domain with total epochs of $100$ and batch size of $64$.

\paratitle{Diffusion Training on Source Domain.}
In our graph score networks, we employ a configuration of $4$ message-passing layers equipped with $8$ attention heads. For the estimation of the final score, we utilize an MLP with $2$ hidden layers. The training of these models is accomplished with the use of an Adam optimizer, maintaining a constant learning rate of $2e-5$ and batch size of $128$. To enhance stability in the parameter updating process, we also implement the exponential moving average~(EMA) methodology, applying a momentum of $0.9999$. 
We employ the variance preserve SDE in this work. We utilize the Euler-Maruyama method, setting $\delta t$ at $0.001$, with $1000$ discretization steps for graph sampling during the reconstruction phase.


\paratitle{Source-free Adaptation on Target Domain.}
During the target domain adaptation, we initially perform domain alignment on the target data using an off-the-shelf diffusion model. 
The reconstruction starting point $t_{recon}$ is default set at $0.1$. That means we first add noise from $t=0$ to $t=t_{recon}$, followed by a denoising process back to $t=0$ for domain transfer. Subsequently, we employ unbiased pseudo-labeling with curriculum learning and consistency learning with graph jigsaw for the domain adaptation.
In unbiased pseudo-labeling learning, the initial setting for the pseudo-label threshold $\alpha(0)$ is $0.95$. In curriculum learning, the confidence threshold $\alpha(e)$ increases linearly with the epoch $e$ from $\alpha(0)$ to $0.99$. In the graph jigsaw, we dynamically partition graphs into communities using the Louvain algorithm. Following this partition, subgraphs are exchanged among confident and unconfident graphs, and the exchanged subgraphs inherit the edges connecting them from the original graph.

\begin{figure*}[t]
    \centering
    \includegraphics[width=0.9\textwidth]{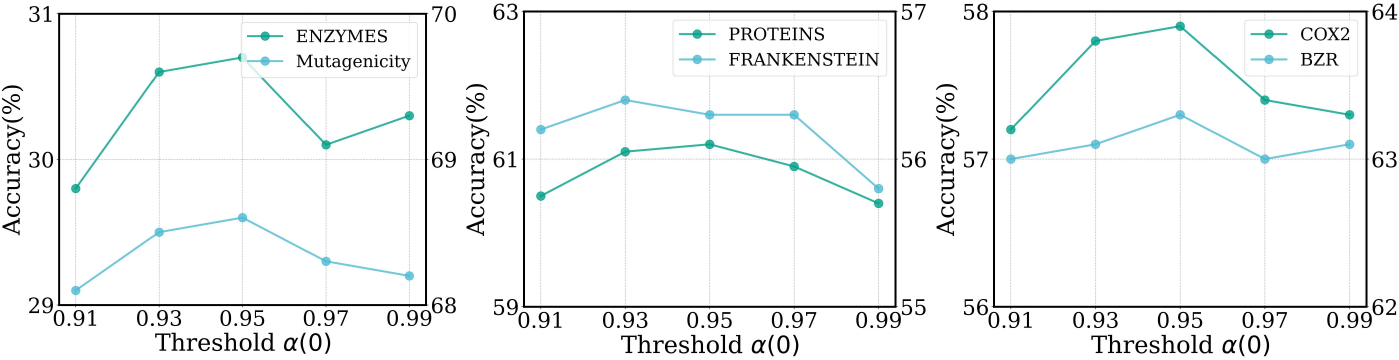}
    \caption{Sensitivity analysis of the pseudo-label threshold. }
    \label{fig:sup-analysis-threshold}
\end{figure*}

\begin{figure*}[t]
    \centering
    \includegraphics[width=0.9\textwidth]{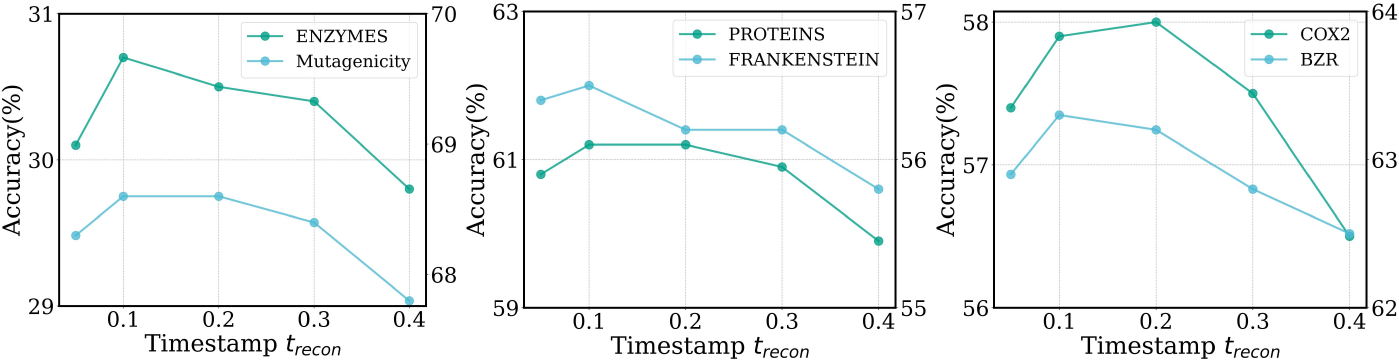}
    \caption{Sensitivity analysis of the reconstruction timestamp. }
    \label{fig:sup-analysis-timestamp}
\end{figure*}

\subsection{Performance Comparison}

Table~\ref{tab::1},~\ref{tab::2}, ~\ref{tab::3} and~\ref{tab::4} present the performance of different methods. We observe the following: 

1) Source-free domain adaptation presents a challenging task, as the inferior accuracy in the target domain, highlights the necessity of investigating this problem. Both domain shifts and the unavailability of source data impose constraints on the model's capabilities, rendering prior research inapplicable.

2) Semi-supervised methods~(\eg, InfoGraph) perform generally better than source-only methods. Semi-supervised methods can utilize both labeled data~(source domain) and unlabeled data~(target domain). Note that in practical scenarios, it is often not feasible to access the source data, as only off-the-shelf models are available. Nevertheless, semi-supervised methods exhibit poor stability and performance, as the high standard deviation in experiments. This is attributed to the absence of consideration for domain shift between the source domain and the target domain.

3) Source-free method PLUE performs better than the other approaches, especially in tackling the challenges posed by more complex domain adaptation tasks. 
Despite PLUE being the state-of-the-art method for source-free domain adaptation in image classification tasks, it is noteworthy that this method is not devised for graph data and substantial domain shifts. Consequently, there still remains a need for further improvements.

4) \method{} has yielded significant improvements, both within source-free domain adaptation on the split sub-datasets and cross-datasets. Notably, \method{} demonstrates an average improvement of $5.3\%$ in cross-dataset experiments when compared to the best-performing compared baseline methods. Furthermore, GALA exhibits improvements, especially in cases of poorly performing domain adaptation sub-tasks.

The improvements in \method{} can be attributed to two primary factors: data transformation and domain transfer. 1) Diffusion-based data transformation allows target data to transform to the source domain while preserving semantics. This mitigates the adverse effects of domain shift and facilitates better predictions in the source domain. 2) The integration of graph jigsaw and proposed pseudo-label methods significantly helps the model learn more robust representations during the adaptation process, thereby enhancing the adaptation of the model.

\subsection{Visualization}
In order to further analyze the effectiveness of \method{}, we investigate the two newly proposed modules, \ie~the graph diffusion and the graph jigsaw. 

\paratitle{Graph Diffusion.} To validate the effects of diffusion-based data adaptation, we performed visual verification on the target domain data in ENZYMES and PROTEINS. As shown in Figure~\ref{fig:sup-diff-enzymes} and Figure~\ref{fig:sup-diff-proteins}, diffusion-based data adaptation makes the target graph more compatible with the source model. Graph diffusion enables domain transfer for graph data and the generation of source-style graphs.

Graph Diffusion modifies the graph structure of a target domain by adding or removing edges between nodes while preserving the original semantic information of the graph. This allows the target graph to become more similar to the source domain, facilitating more accurate predictions by models trained on the source domain.

\paratitle{Graph Jigsaw.} We visualize the graph jigsaw process. As in Figure~\ref{fig:sup-vis-jigsaw}, it can be observed that our proposed graph jigsaw is able to adaptive select subgraphs and perform exchanges between confident and unconfident graphs. In graph jigsaw, we randomly chose a subgraph $\hat{G}_{j,2}^c$ from each confident graph $\hat{G}_j^c$, and exchanged with an unconfident graph $\hat{G}_k^u$ to generate an augmented graph $\hat{G}_j^c$. The visualization elucidates the effective execution of graph jigsaw, like a jigsaw puzzle game, by employing the copy-and-paste mechanism with subgraphs. 

The exchange of subgraphs and establishment of connections generate new graph data, which helps explore data samples. With graph jigsaw, the model learns more robust representations by consistency learning between the original graphs and post-exchange graphs, therefore enhancing the generalization ability on the target domain.

\begin{figure}[t] 
    \centering 
    \begin{subfigure}[b]{0.48\columnwidth} 
        \includegraphics[width=\textwidth]{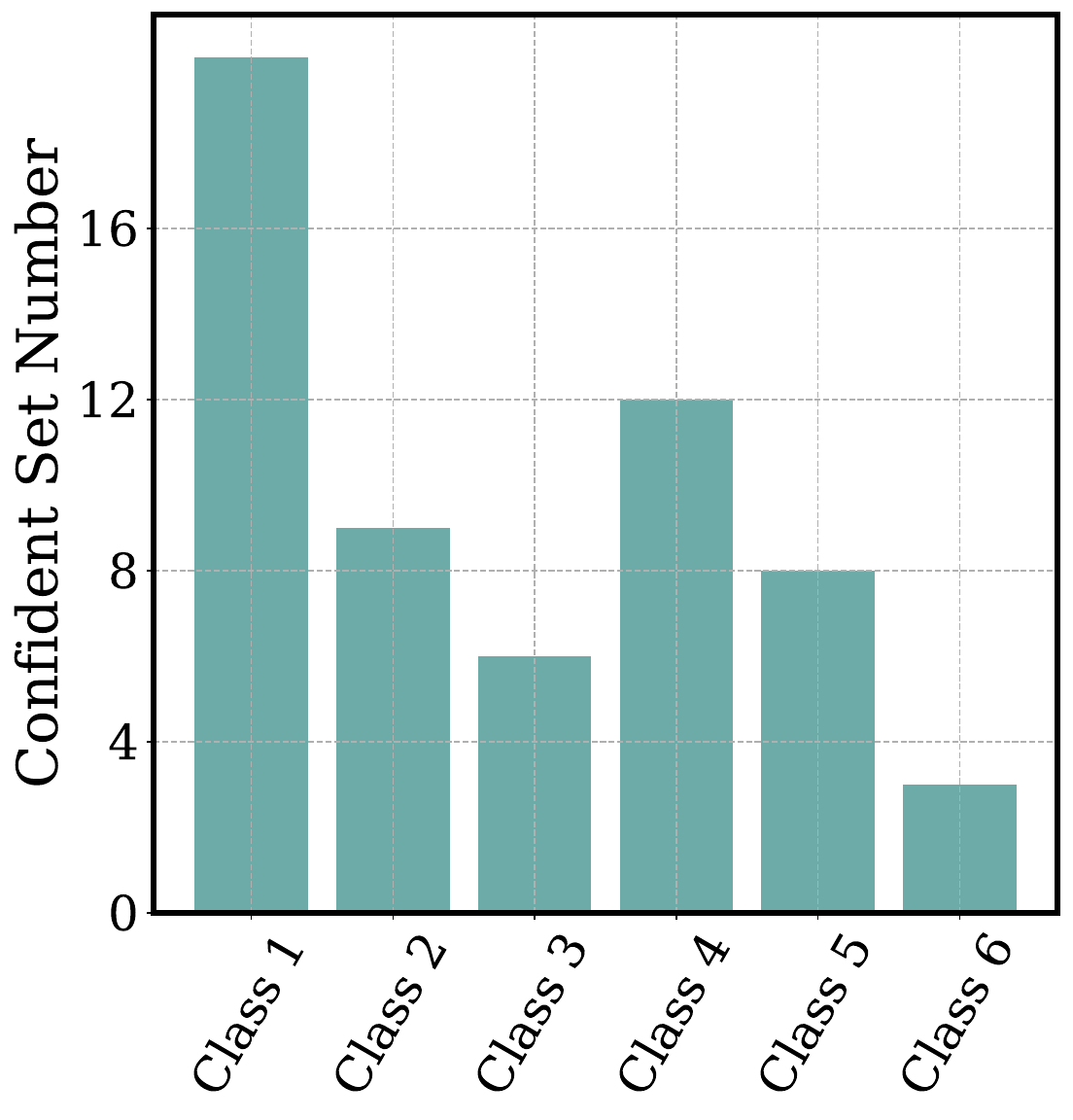}
        \caption{\fix{Pseudo-labeling}}
    \end{subfigure}
    \hfill 
    \begin{subfigure}[b]{0.48\columnwidth}
        \includegraphics[width=\textwidth]{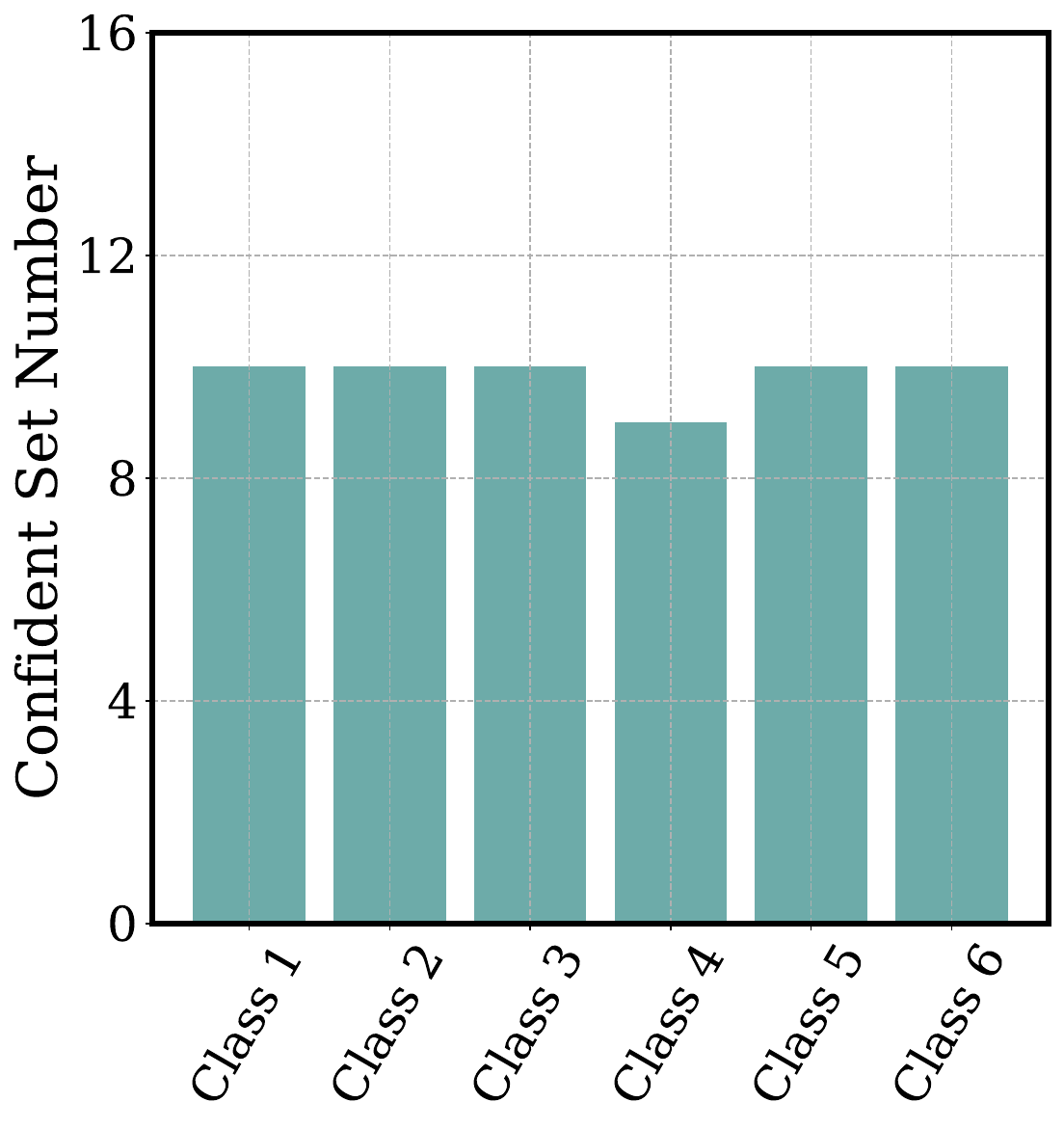} 
        \caption{\fix{Unbiased Pseudo-labeling}}
    \end{subfigure}
    \caption{\fix{Validation on ENZYMES. Pseudo-labeling with a fixed threshold can be biased towards easy classes~(left). Our unbiased pseudo-labeling can provide more unbiased pseudo-labeling~(right).}}
    \label{fig:unbiased-pseudo-labeling} 
\end{figure}

\begin{figure}[t] 
    \centering 
        \includegraphics[width=0.9\columnwidth]{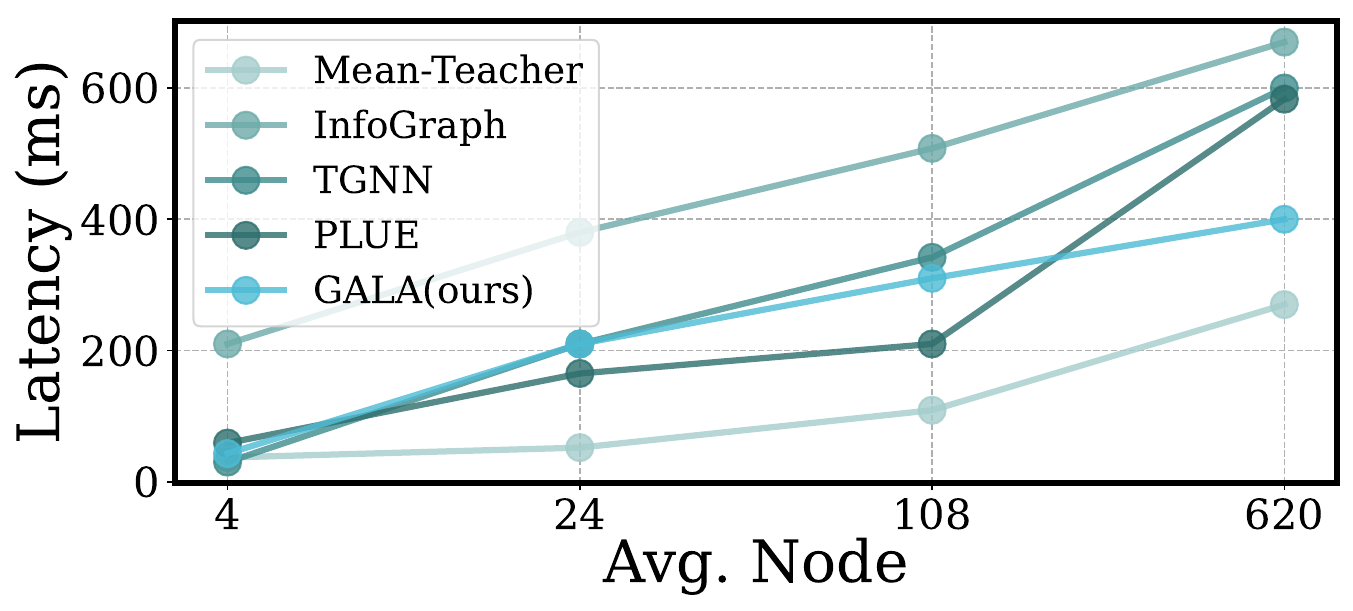}
    \caption{\fix{Scalability analysis. We evaluate the adaptation latency for a single batch across different graph sizes and methods.}}
    \label{fig:scalability} 
\end{figure}

\subsection{Ablation Study}

We investigate the effectiveness of \method{} in three aspects: the graph diffusion model for adaptation~(GDM), unbiased pseudo-labeling with curriculum learning~(UPCL), and consistency learning with graph jigsaw~(CLGJ). The average classification results on two datasets are summarized in Table~\ref{tab::abl} and we have four results. \textit{Firstly}, it can be observed that the whole model $M_4$ achieves the best performance. 
As such, adaptation is required from both data and models. The collaborative interaction of these three modules can enhance the effectiveness.
\textit{Secondly}, different components can play distinct roles. GDM appears to be a notably effective component~($M_1$), suggesting that data adaptation serves a crucial role in handling distribution shifts, the significant challenge in source-free domain adaptation. 
\textit{Thirdly}, the role of UPCL varies in magnitude, depending on the bias present in pseudo-labels. 
plays a more important role in ENZYMES~($M_2$), indicating a more serious biased of pseudo-labels. Nonetheless, consistently, they all play a significant role in the overall result of source-free domain adaptation.
\textit{Fourthly}, we can observe that CLGJ shows relatively consistent efficacy across both datasets~($M_3$), which represents the gain from consistency learning with graph jigsaw.

\fix{In addition, to empirically analyze the effectiveness of UPCL, we conducted studies on the ENZYMES~(E0$\rightarrow$E3), focusing on the confidence set selection in the target domain. As shown in Figure~\ref{fig:unbiased-pseudo-labeling}, the traditional pseudo-labeling methods can generate biased confidence sets that are skewed toward easier categories, which negatively affects adaptation in the target domain. In contrast, our approach to unbiased pseudo-labeling shows significant effectiveness in preventing such biases.}

\subsection{\fix{Scalability Analysis}}

\fix{Following to the experimental setup in Section~\ref{sec:imple}, we analyze the scalability of our proposed \method{} using the PROTEINS dataset, as shown in Figure~\ref{fig:scalability}. The experiments are conducted on a single RTX 3090 GPU, where we sample data from different scales of graph data and report the adaptation latency per batch. The results show that our method exhibits high efficiency as graph size increases, outperforming other methods. This highlights the ability of our method to achieve efficiency and high performance when handling large-scale graph data.}

\subsection{Sensitivity Analysis}

\paratitle{Analysis of pseudo-label threshold.} 
We first vary the initial pseudo-label threshold $\alpha(0)$. Combined with class-wise accuracy and training epochs, $\alpha(0)$ determines the confidence threshold. Its impact on final average accuracy is investigated in Figure~\ref{fig:sup-analysis-threshold}.
We test $\alpha(0)$ from $0.91$ to $0.99$ with uniform sampling intervals..
As $\alpha(0)$ from $0.91$ to $0.95$, the final accuracy improves on all datasets. This indicates that relatively clean pseudo-labels benefit model adaptation. However, when $\alpha(0)$ is further increased to $0.99$, there is a decrease in accuracy. This suggests that an excessively strict pseudo-label threshold can limit the model's ability to learn from the target domain.
Therefore, in order to make a balance between these two considerations, we take $\alpha(0)=0.95$ as the default value, enabling us to maximize the utilization of the information within the target domain.

\paratitle{Analysis of reconstruction timestamp.} 
We investigated the impact of reconstruction timestamp $t_{recon}$, which is the end point of forward SDE and the start point of reverse SDE.
$t_{recon}$ influence semantic information preservation during reconstruction. We vary $t_{recon}$ in $[0.05, 0.1, 0.2, 0.3, 0.4]$ to investigate the potential effects arising from diffusion reconstruction. 
As shown in Figure~\ref{fig:sup-analysis-timestamp}, the optimal average accuracy is achieved when the $t_{recon}$ is set to $0.1$.
As $t_{recon}$ increases, the overall accuracy rises, which shows the necessity of employing data domain transfer based on graph diffusion for the target graph data. However, as $t_{recon}$ continues to increase, we observe varying degrees of decline in overall accuracy. This suggests that an excessive alteration of the information in the target graph data can generate side effects, impacting the original semantic information within the graph.
Insufficient diffusion leads to inadequate data adaptation, while excessive diffusion hinders the preservation of graph semantics. They both result in an adverse impact on accuracy.
\fix{This analysis reveals that transforming target graphs into source-style graphs carries a potential risk of information loss due to over-reconstruction. \method{} can transfer graph semantics with limited information loss, which can improve the graph domain adaptation performance.}

\section{Conclusion and Future Works}

This paper studies source-free domain adaptation, a practical scenario in graph domain adaptation, where the source data is inaccessible during adaptation, resulting in the domain shift and label scarcity of the target domain. We propose \method{} to address these problems.
\method{} utilizes graph score-based diffusion for domain alignment, unbiased pseudo-labeling with curriculum learning, and consistency learning with graph jigsaw. Extensive experiments demonstrate the effectiveness of \method{}. 

\paratitle{Future work.} \fix{Our methods could be applied to heterogeneous graphs for complex real-world environments in future work. In heterogeneous graphs, capturing intricate structural and semantic information remains a challenge. In addition, the diversity of node and edge types in heterogeneous graphs, each potentially carrying different types of information, increases the risk of information loss during the reconstruction process. These challenges deserve further investigation.}

\bibliographystyle{IEEEtran}
\bibliography{rec}

\end{document}